\documentclass[11pt]{article}
\usepackage[letterpaper,margin=1in]{geometry}
\usepackage{amsmath,amssymb,amsthm}
\usepackage{graphicx,booktabs,array,longtable}
\usepackage[numbers,sort&compress]{natbib}
\usepackage[hidelinks]{hyperref}
\usepackage{microtype}
\hypersetup{pdftitle={Good Views Are Not Enough: Global Counterexamples for Gaussian-Regularized JEPAs},pdfauthor={Anran Xu}}
\newtheorem{theorem}{Theorem}
\newtheorem{lemma}[theorem]{Lemma}

\newtheorem{corollary}[theorem]{Corollary}
\theoremstyle{remark}

\newcommand{\E}{\mathbb E}
\newcommand{\R}{\mathbb R}
\newcommand{\Var}{\operatorname{Var}}
\newcommand{\tr}{\operatorname{tr}}

\newcommand{\norm}[1]{\left\|#1\right\|}
\newcommand{\cf}{\varphi}
\newcommand{\eps}{\varepsilon}
\newcommand{\sigmoid}{\operatorname{sigmoid}}

\makeatletter
\newenvironment{figurehere}{\par\medskip\noindent\begin{minipage}{\linewidth}\centering\def\@captype{figure}}{\end{minipage}\par\medskip}
\newenvironment{tablehere}{\par\medskip\noindent\begin{minipage}{\linewidth}\centering\def\@captype{table}}{\end{minipage}\par\medskip}
\makeatother
\title{Good Views Are Not Enough:\\
Global Counterexamples for Gaussian-Regularized JEPAs}
\author{Anran Xu\\School of Automation and Intelligence\\Beijing Jiaotong University}
\date{November 9, 2026}
\begin{document}
\maketitle
\begin{abstract}
Good views do not suffice to guarantee scene invariance or uniqueness up
to rotation at the global optima of Gaussian-regularized JEPAs.
We give globally certified counterexamples showing that the uniqueness
premise of the Gaussian extension in \emph{Good Views} fails for specified
soft objectives. Gaussian marginal matching can reward retaining
independently resampled observation variation, and this benefit can exceed
its prediction cost. In a four-state model, comparison against the entire
scene-only class proves that every global minimizer splits a scene and
admits an equal-value partner outside its rotation orbit.
In a 512-dimensional model with a complete scene basis, a spectral gap,
and exact population whitening, pooled Epps--Pulley regularization still
forces every global minimizer to retain nuisance and yields positive
cross-modal alignment error even after the best rotation when modality
nuisances are independent. Better scene alignment can also accompany worse
prediction. With matched 512-label readout budgets and 1\% observation
noise, whitened prediction JEPA aligns better to the scene than next-token
prediction but incurs 0.2872 nats/token higher linear-readout test NLL.
A Bayes-risk decomposition distinguishes information lost through the
observation channel from the fitted readout's difficulty using retained
information. These results connect three determinants of learned
representations: what views share, what regularization favors, and what
a prediction task can use.
\end{abstract}

\section{Introduction}
\label{sec:intro}
What views share, what an objective retains, and what a task needs are
distinct questions. For a picture and its description, the shared content
may be a scene; lighting, wording, or other observation details can vary.
Good views specify the shared relation. The regularized objective determines
which variation enters the representation, and the prediction task determines
the value of that retained content.

Prediction penalizes variation across positive observations, while Gaussian
matching can reward retaining it. With finitely many scenes, a deterministic
scene-only encoder produces a finite atomic distribution.
Independently resampled observation variation supplies
additional atoms and can improve the Gaussian fit. When that improvement
exceeds its prediction cost, \emph{every} global optimum retains variation
that the views do not share.

Good Views explains cross-modal agreement through a shared world objective
\citep{goodviews}. We examine how Gaussian marginal regularization changes
the content selected by that objective. Our constructions compare the
combined loss over the entire scene-only class, retaining a scene spectral
gap and, in the high-dimensional model, exact whitening.

Retained variation can also be useful predictive context. Scene alignment
rewards ignoring wording differences between descriptions of the same scene;
token prediction can require them. A separate four-state task examines
this distinction with frozen encoders, matched readout labels, observation
noise, and head architectures.

We establish three results.

\noindent\textbf{Global failure of invariance and uniqueness.}
A four-state construction forces every global optimum to split a scene
and yields attained, equal-value minimizers outside a single sign orbit.

\noindent\textbf{Nuisance retention under exact whitening.}
With a complete scene basis, a spectral gap, and exact whitening in 512
dimensions, pooled Gaussian regularization forces nuisance into every global
optimum and yields a positive best-rotation cross-modal alignment error
under independent modality nuisances.

\noindent\textbf{Different rankings for alignment and prediction.}
A matched readout study shows how scene alignment and predictive risk
respond differently to retained observation variation, separating
observation loss from the fitted head's gap.

\section{Related work}
\label{sec:related}
\subsection{Views, invariant content, and identifiability}
Good Views connects cross-modal alignment to a common world
optimization problem \citep{goodviews}. Its Theorem 1 assumes that
the shared objective has a unique optimum up to rotation.
Proposition 1's whitened prediction analysis establishes this property
under exact whitening and a spectral gap; Proposition 3 derives independent
nuisance elimination under the same objective. For the Gaussian
extension, uniqueness remains a premise. The distinction opens a
concrete optimization question: how does a Gaussian penalty interact
with the invariant spectral solution? Our constructions answer it
for specified soft objectives by proving full-class exclusion,
attainment, and the resulting structure of global minimizers.

Content--style theory gives a complementary route from views to
invariant content. In smooth continuous latent models, alignment
combined with differential entropy can isolate content up to an
invertible map \citep{contentstyle}. A content-only encoder can
simultaneously achieve perfect alignment and the desired output entropy.
Structured augmentations further use known patterns of shared factors
to recover separate representation blocks \citep{structured}.
These results highlight compatibility between the view relation,
the representation class, and the output objective. Our finite scene
models expose the competing case: a scene-only marginal has a
quantified Gaussian discrepancy, and independent observation variation
reduces that discrepancy enough to change the joint optimum.

Distribution-preserving ambiguities are a classical obstacle in
nonlinear identifiability \citep{pitfalls}. The pair relation adds
structure that can remove some marginal symmetries. We therefore analyze
both terms of the objective: a uniform bound first forces a split
within a scene, and a loss-preserving swap then supplies a second
minimizer outside the rotation orbit. This joins a symmetry argument
to an attained global optimum and a certified comparison over the
whole invariant class.

\subsection{Gaussian regularization and representation content}
LeJEPA introduces sliced characteristic-function matching to a
standard Gaussian \citep{lejepa}. KerJEPA relates the Epps--Pulley
statistic to Gaussian-kernel MMD and develops its estimators and
finite-batch corrections \citep{kerjepa}. Those tools make the
regularizer amenable to exact analysis. We use their discrepancy
and expectation identities to compare scene-only and split
representations at specified frequencies, batch sizes, and weights.

Positive Gaussian world-model results combine marginal constraints
with transition structure. \emph{When Does LeJEPA Learn a World Model?}
uses Gaussian states and Ornstein--Uhlenbeck transitions to identify
orthogonal linear representations \citep{worldmodel}. Controlled-world
identifiability adds sufficiently expressive action-conditioned
predictors and spectral conditions \citep{controlled}. The anisotropic
JEPA analysis proves uniqueness of the encoder Gram matrix within a
linear Gaussian class and studies how representation geometry affects
planning \citep{anisotropic}. Together these works explain when
Gaussian matching and the predictive spectral solution are compatible.
Our 512-dimensional construction fixes a complete scene basis and exact
whitening, then isolates how the additional pooled EP term selects
independently resampled variation.

Gaussian output also has an information-bottleneck motivation
\citep{normal}, while LRC-JEPA distinguishes dynamics from residual
context and explains why aggregate Gaussian matching alone does
not determine minimality \citep{lrcjepa}. Slow trajectory noise provides
another mechanism for retaining unwanted variation in predictive
learning \citep{contrast}. Here the nuisance is independently redrawn
at positive endpoints. The certified global gap attributes its retention
to marginal matching and gives a positive lower bound for every optimum
of the specified whitened soft objective.

\subsection{Representation geometry and predictive usefulness}
Good Views defines nuisance relative to alignment and interprets
next-token prediction through an algebraic decomposition followed
by a heuristic view model \citep{goodviews}. This makes the evaluation
task central to deciding which variation matters. Empirical InfoNCE
work already reports tradeoffs between identification proxies and
classification performance, with the outcome depending on the
augmentations and data-generating process \citep{infonce}. Anisotropic
JEPA likewise connects a planning gap to task geometry in an invertible
representation \citep{anisotropic}.

Our readout study gives a controlled comparison with known latent
states and token probabilities. Frozen encoders share finite label
budgets, heads, and noisy observation channels. Conditional Bayes risk
\citep{cover} separates information available through that channel
from the fitted head's remaining gap. This exposes how better scene
alignment can accompany worse prediction and how head capacity and
task dependence shape that ordering.

\section{Views and Gaussian regularization}
\label{sec:objectives}
Let \(S\) denote the scene and \(U\) observation variation. A view is
an injective rendering of \((S,U)\). Good views resample \(U\)
independently at positive endpoints; the positive relation lies in the
scenes. A scene-only encoder \(h\) satisfies
\(h(S,U)=r(S)\). Thus changing \(U\) within a scene has no effect on
its representation. We call \(U\) \emph{nuisance} relative to this
view relation; whether it is useful for prediction depends on the task.

The optimization has two preferences. The prediction term \(P\)
penalizes squared distances to the global-view center. The Gaussian
term compares the output characteristic function
\(\cf(t)=\E e^{it h}\) with \(g(t)=e^{-t^2/2}\), using weighted squared
differences. For \(m\) observations, the statistic is
\(\widehat R=m\sum_j\omega_j|\widehat\cf_m(t_j)-g_j|^2\).
Its expectation depends on whether views are processed separately
or pooled. Each theorem specifies that choice and its coefficient;
Appendix~\ref{app:objectives} gives the normalization formulas.

\begin{figure}[ht]
\centering
\includegraphics[width=\linewidth,trim=0 4pt 0 4pt,clip]{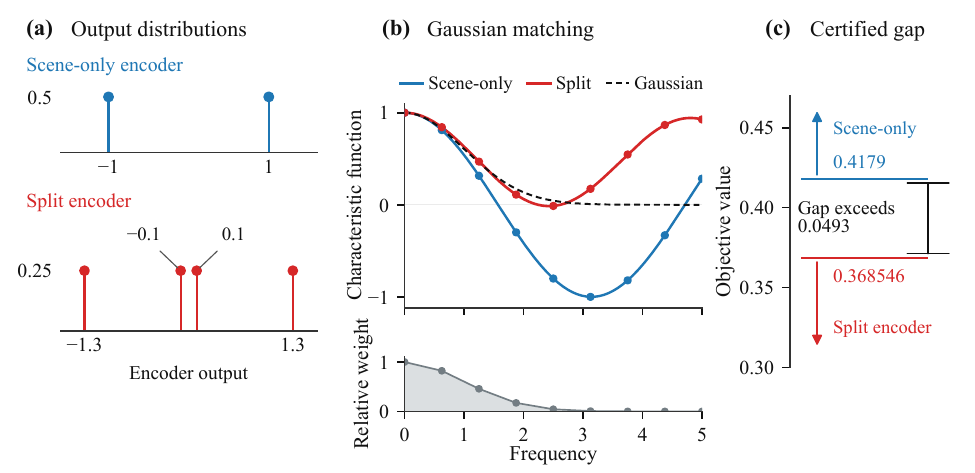}
\caption{Gaussian matching and scene splitting.
Panels (a--b) illustrate a scene-only encoder and the four-atom comparator;
panel (c) compares the comparator's objective with a certified lower bound
for the entire scene-only class. The comparator maps the positive scene to
\(\{0.1,1.3\}\) and the negative scene to \(\{-0.1,-1.3\}\).}
\label{fig:mechanism}
\end{figure}

Figure~\ref{fig:mechanism} illustrates the marginal-fit competition and
the full-class objective gap. Dots mark the nonnegative half of the
17-node grid; negative nodes follow by even symmetry.
The bounds use 1024 iid samples per view and mixing weight 0.02.

\section{Gaussian matching breaks scene invariance and uniqueness}
\label{sec:scalar}
Two scenes with two observations each give at most two output values
to any scene-only encoder. A uniform positive Gaussian-discrepancy bound
covers this entire class; a four-atom marginal improves matching enough
to outweigh the cost of within-scene variation (Figure~\ref{fig:mechanism}).

Index the observations by
\((S,U)=(1,1),(1,-1),(-1,1),(-1,-1)\), with uniform probability \(1/4\).
Set \(s=(1,1,-1,-1)\) and
\[
 \pi_{ij}=\frac{1+s_i s_j/2}{16},\qquad
 P_\pi(x)=\frac14\sum_{i,j}\pi_{ij}(x_i-x_j)^2.
\]
All positive-pair entries are positive. The transition eigenvalues
are \(1,1/2,0,0\); the scene direction has a positive gap over the
independent nuisance directions. Work on the centered domain
\(X_0=\{x\in\R^4:\sum_i x_i=0\}\).

For the 17 signed nodes \(t_j=5j/8\), \(-8\leq j\leq8\), use
\(\omega_j=(5/8)c_j e^{-t_j^2/2}\), with endpoint factors
\(c_{\pm8}=1/2\) and all others one. Let \(D_{\rm gr}\) be the
weighted characteristic-function discrepancy, and \(R_{1024,\rm gr}\)
the exact expectation of the EP term on a per-view iid batch of 1024.
EP is averaged across the two views. The chosen soft objective is
\begin{equation}
 F_\pi(x)=P_\pi(x)+\frac1{49}R_{1024,\rm gr}(\mu_x),\qquad
 \mu_x=\tfrac14\sum_i\delta_{x_i}.
 \label{eq:scalar_primary}
\end{equation}
This is the mixed coefficient \(\lambda=1/50\) after dividing the
loss by \(1-\lambda\).

\begin{samepage}
\begin{theorem}[Attained nonorthogonal scalar minimizers]
\label{thm:scalar_primary}
The objective \eqref{eq:scalar_primary} attains its minimum on \(X_0\).
Every global minimizer takes distinct values within at least one
scene. For every such minimizer \(x\), exchanging the two observations
of one scene gives an equal-value global minimizer outside
\(\{x,-x\}\).
\end{theorem}
\end{samepage}

\paragraph{Why the conclusion is global.}
The entire scene-only class is strictly suboptimal.
Every scene-only vector \((b,b,c,c)\) has
\(D_{\rm gr}>1/50\). The iid expectation identity
\(R_N=ND+A_\nu-J_\nu\), with \(0\leq J_\nu\leq A_\nu\),
therefore gives \(F_\pi>1024/2450>0.4179\).
The rational comparator
\[
 w=(1/10,13/10,-1/10,-13/10)
\]
is centered, has \(P_\pi(w)=121/400\), and satisfies
\(D_{\rm gr}(\mu_w)<0.00216239\).
The sharper interval evaluation gives
\(F_\pi(w)<0.368546\). The gap exceeds \(0.0493\), excluding the whole scene-only class.
The lower bound uses a rational interval cover of the complete
two-atom phase domain.

Three further steps turn exclusion into nonuniqueness.
First, \(P_\pi\geq\Var(x)/4\) makes the centered objective coercive,
so a minimum exists. Second, any nonconstant optimum has nonzero
scene correlation: rearranging its values otherwise lowers prediction
without changing its marginal law. Third, a split-group swap preserves
both terms of the loss; if it were an overall sign reversal, the
scene correlation would vanish. Appendix~\ref{app:scalar} supplies the
full proof, analytic-integral version, and other sufficient weights.
A three-feature fixed ReLU network realizes every four-state encoder
(Appendix~\ref{app:realization}).

The construction links a quantitative marginal-fit gap to the
geometry of the global solution set. Appendix~\ref{app:stability}
extends the mechanism to separated near-optimal partners when the
pair symmetry is perturbed by at most
\(5\times10^{-6}\) in total entrywise distance. Their sign-orbit
distance exceeds \(0.025\), and excess loss is below \(3\times10^{-5}\).

\section{Gaussian matching forces nuisance under exact whitening}
\label{sec:highdim}
The scalar construction establishes the content-selection mechanism and
its consequence for uniqueness. We next place the same competition inside
an exactly whitened model with a complete scene basis. Prediction alone
selects a scene-only representation; an additional pooled EP term makes
\emph{every} scene-only encoder strictly suboptimal.

Let \(S\) have 513 values, with \(p_1=p_2=1/4\) and
\(p_s=1/1022\) for the other 511 scenes. Independently draw
\(U\) uniformly on \(\{-1,1\}^{512}\). All eight views keep the
same scene and independently redraw \(U\). A complete centered
orthonormal scene basis \(r(S)\in\R^{512}\) supplies an injective
rendering \((r(S),U)\). The view transition has 512 nonconstant
scene eigenvalues one and all nuisance eigenvalues zero.

Optimize all finite-support encoders \(h(S,U)\in\R^{512}\) under
\(\E h=0,\ \E hh^\top=I_{512}\). Define the nuisance fraction
\[
 u(h)=\frac1{512}\E\norm{h-\E[h\mid S]}^2.
\]
With \(G=2\) global views among \(V=8\) total views, the
coordinate-mean center loss is \(P=5u/4\).
Pool \(N=256\) examples and their views into \(M=2048\) outputs.
The EP statistic uses nodes \(t_j=3j/16\), \(0\leq j\leq16\),
with \(\omega_j=(3/16)d_jg_j\), \(d_0=d_{16}=1\), and
\(d_j=2\) at interior nodes.
Average directions \(\theta\) \emph{exactly} over
\(\mathbb S^{511}\) and the statistic over the batch:
\begin{equation}
 F(h)=\frac54u(h)+\frac1{50}\bar R(h),\qquad
 \bar R=\E_{\mathcal B,\theta}\widehat R_{\mathcal B,\theta}.
 \label{eq:high_primary}
\end{equation}
The direct coefficient follows \texttt{stable-pretraining} 0.1.8
\citep{stable}; the domain additionally imposes exact population whitening.

\begin{theorem}[All whitened soft-objective optima retain nuisance]
\label{thm:high_primary}
The objective \eqref{eq:high_primary} attains a minimum on this
domain, and every global minimizer satisfies \(u(h)>0.00015\).
Two global minimizers with independent modality nuisances on the
same scene have best-rotation coordinate-mean MSE greater than
\(0.0003\). Both statements also hold in the linear class
\(h_W=W(r,U)\), \(WW^\top=I_{512}\).
\end{theorem}

\begin{figure}[ht]
\centering
\includegraphics[width=\linewidth,trim=0 4pt 0 4pt,clip]{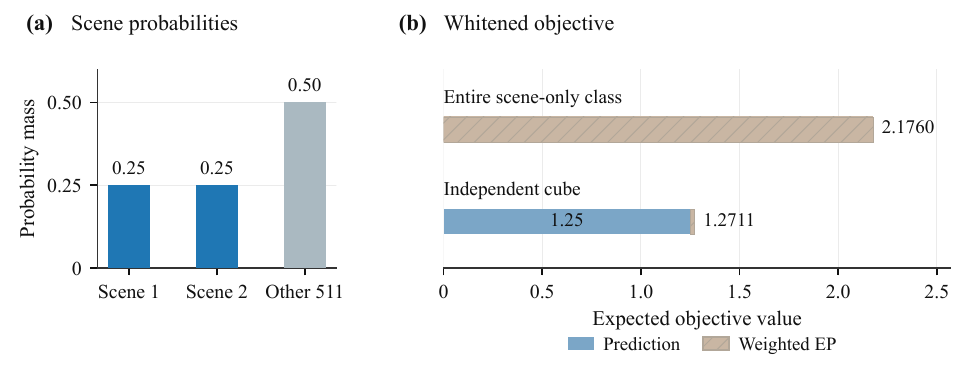}
\caption{Whitened global comparison.}
\label{fig:whitened}
\end{figure}

\paragraph{Why one comparison excludes every scene-only solution.}
The comparison excludes every scene-only whitened encoder: all such
encoders share the same direction-averaged EP value, and a feasible
nuisance encoder has strictly lower total loss.
The complete scene basis satisfies
\(r(s)^\top r(s')=\mathbf1_{s=s'}/p_s-1\).
Every scene-only whitened encoder is \(Qr\), \(Q\in O(512)\).
Its two common-scene norms are only \(\sqrt3\), whereas the other
norms are \(\sqrt{1021}\). Those constraints on half the marginal
mass give the same EP value uniformly over rotation.
Figure~\ref{fig:whitened} separates prediction from EP weighted by \(0.02\).
The scene-only class has prediction loss zero, EP \(108.798853\),
and full objective \(2.175977\). The whitened independent cube \(h=U\)
has prediction loss \(1.25\), EP \(1.053856\), and objective \(1.271077\).
These are rounded certified expectations under exact sphere and batch
averaging, with strict bounds \(F_s>2.1759\) and \(F_U<1.2711\).
The cube's EP gain outweighs its prediction cost.

The strict gap exceeds \(0.9\). For a general encoder, projecting
onto its conditional scene mean and taking a polar factor produces
a scene-only rotation with
\(F_s\leq F(h)+6000u(h)\). Compactness gives attainment, and the
cube upper bound then implies
\(u>0.9/6000=0.00015\).
Conditional independence converts this fraction into the stated
cross-modal MSE bound. Appendix~\ref{app:highdim} gives the full
moment, sphere-integral, and polar-decomposition argument.

This is a content-selection effect under full scene realizability:
the marginal-fit gain exceeds the prediction cost throughout the
scene-only class. Appendix~\ref{app:batch} extends the comparison
to a single random batch with exact direction averaging:
with probability at least \(0.828\), one realized batch excludes all
scene-only optima, and linear optima have \(u>10^{-12}\).

\section{Better scene alignment can accompany worse prediction}
\label{sec:readout}
Having shown that regularization can determine which observation variation
is retained, we now ask what that variation means for prediction.
Variation that is nuisance for scene alignment can be useful predictive
context. We study this question in a separate four-state task.
Let \((S,U)\) be uniform on four sign states and let a binary token obey
\begin{equation}
 \Pr(Y=1\mid S,U)=\sigmoid(S+bU).
 \label{eq:token}
\end{equation}
The main task takes \(b=2\); the negative control takes \(b=0\).
Here \(U\) is alignment nuisance and, for \(b=2\), useful predictive
context.

\paragraph{Matched test of learned representations.}
We compare prediction with whitening (whitened prediction JEPA),
prediction with pooled EP (soft EP JEPA), and prediction with both
whitening and pooled EP (whitening-plus-EP JEPA) with next-token
prediction (NTP). Each objective is trained with five seeds using the same
four-state ReLU parameterization.
Frozen, standardized encoders provide observations
\(Z=z+\sigma\xi\), with \(\xi\sim\mathcal N(0,1)\), to matched linear
logistic or 16-hidden-unit tanh MLP heads. Three independent data repeats
pair states, labels, and noise across methods. The primary comparison
uses 512 training labels, \(\sigma=0.01\), and a linear head.
Test data select no parameters; training steps, validation budgets,
test sizes, controls, and fitting rules appear in
Appendices~\ref{app:protocol}--\ref{app:results}.
Clean scene alignment is \(\min_{q\in\{-1,1\}}\E(z-qS)^2\);
token risk is evaluated through the specified noisy channel.

\begin{table}[ht]
\centering
\begin{tabular}{lrr}
\toprule
Primary setting & Whitened prediction JEPA & NTP\\
\midrule
Clean best-sign scene MSE & 0.003530 & 1.130827\\
Linear test NLL (nats/token) & 0.674001 & 0.386775\\
MLP16 test NLL (nats/token) & 0.617590 & 0.388431\\
\bottomrule
\end{tabular}
\caption{Whitened prediction JEPA has lower clean scene MSE and higher
token NLL than NTP under matched readout conditions.
The MLP is a secondary comparison.}
\label{tab:primary}
\end{table}

The primary JEPA-minus-NTP NLL difference is \(0.287226\), with
95\% two-axis paired bootstrap interval \([0.280889,0.293062]\).
All five seed means and all 15 seed/repeat cells are positive.
The MLP difference is \(0.229160\), interval
\([0.115088,0.291289]\). Greater head capacity reduces the average
difference; the risk decomposition below identifies observation loss
as the main source of the whitened-prediction branch's population excess risk.

\begin{figure}[ht]
\centering
\includegraphics[width=\linewidth,trim=0 4pt 0 4pt,clip]{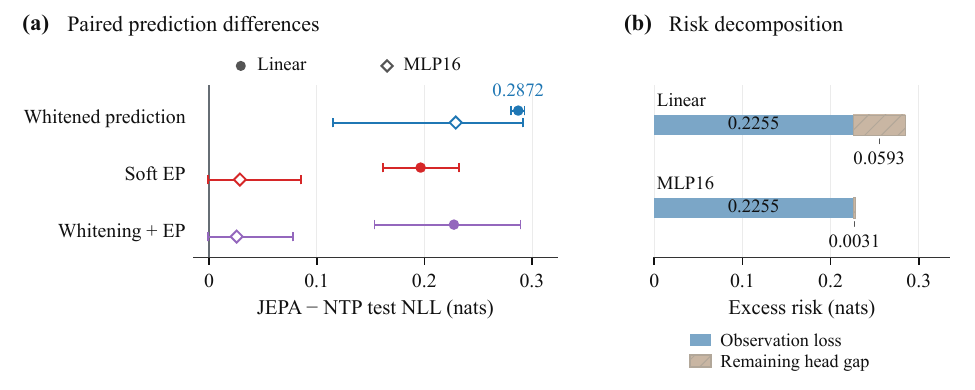}
\caption{Matched prediction differences.
(a) Paired test NLL differences relative to NTP.
(b) Population excess risk relative to the complete-context Bayes risk,
decomposed into observation loss and remaining head gap.}
\label{fig:prediction}
\end{figure}

Figure~\ref{fig:prediction} compares the three JEPA branches with NTP.
Colors identify representations; circles and open diamonds identify
linear and MLP16 heads. The right panel uses numerical Hermite quadrature to decompose
whitened-prediction population excess risk.

\begin{figure}[ht]
\centering
\includegraphics[width=\linewidth,trim=0 4pt 0 4pt,clip]{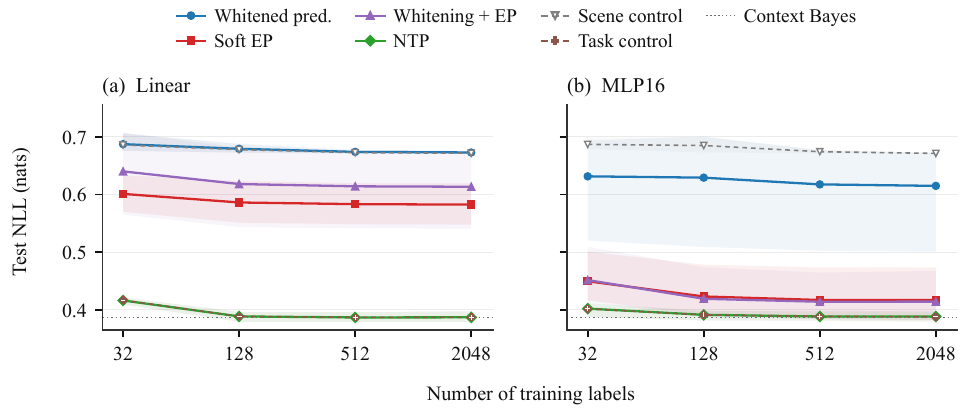}
\caption{Readout learning curves.}
\label{fig:readout_curves}
\end{figure}

Figure~\ref{fig:readout_curves} compares all six representations and
both heads at 1\% noise with 32, 128, 512, and 2048 training labels.
Bands are 95\% two-axis bootstrap intervals over five encoder seeds
and three shared-data repeats. Dashed fixed controls use matched
decoders and data; the dotted line is context Bayes risk,
\(0.386534\) nats/token.
Appendix~\ref{app:results} adds the zero-noise curves.

\begin{samepage}
\paragraph{Observation loss.}
Most of the whitened-prediction JEPA's population excess risk comes from
information unavailable through the noisy observation channel.
Let \(R_{\rm context}\) be Bayes NLL given \((S,U)\),
\(R_{\rm obs}\) Bayes NLL given the actual observation \(Z\),
and \(R_{\rm head}\) the fitted head's population NLL. Then
\begin{equation}
 R_{\rm head}-R_{\rm context}
 =(R_{\rm obs}-R_{\rm context})+(R_{\rm head}-R_{\rm obs}).
 \label{eq:decomp}
\end{equation}
\end{samepage}
The first term is information lost through representation and
observation; the second contains head capacity, finite training,
regularization, and optimization. This is the standard
conditional-entropy decomposition \citep{cover}.
For whitened prediction JEPA, the observation loss averages
\(0.225499\) nats/token. Its linear head gap is \(0.059349\);
the MLP reduces it to \(0.003077\)
(Figure~\ref{fig:prediction}). These head gaps are measured relative
to \emph{the same encoder's observation Bayes risk}.

\begin{figure}[ht]
\centering
\includegraphics[width=\linewidth,trim=0 4pt 0 4pt,clip]{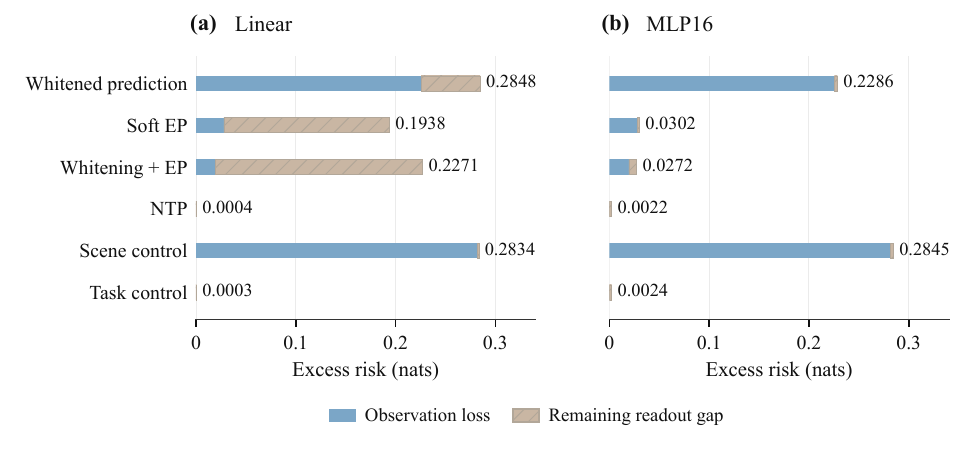}
\caption{Observation loss and readout gaps.}
\label{fig:risk}
\end{figure}

Figure~\ref{fig:risk} decomposes all six representations at 512 labels
and 1\% noise using 128-node Hermite estimates. Solid and hatched bars
separate observation loss and head gap. NTP and the task control have
observation loss below numerical resolution. Head capacity affects the
remaining readout gap; observation Bayes risk gives the irreducible prediction
loss for the specified channel.

At zero noise, all 20 learned encoders have four distinct outputs,
with minimum separation approximately \(7.77\times10^{-16}\).
An ideal exact-value lookup retains full context information, while
positive observation noise makes some differences unstable. One encoder
seed retains more stably readable observation variation than the other
four. Thus exact-value information and information available through
the readout channel differ, including across seeds.

\paragraph{Nonlinear access to retained information.}
For the soft Gaussian branches, a nonlinear readout substantially reduces
the remaining head gap. At the primary setting, the MLP reduces that gap
from \(0.165827\) to \(0.002191\) for soft EP and from \(0.207490\) to
\(0.007651\) for whitening-plus-EP. Their MLP
differences from NTP are \(0.028651\) and \(0.025593\);
both bootstrap intervals cross zero. The zero-noise MLP results also
vary across seeds and label budgets; Appendix~\ref{app:results} gives
the individual reversals. Much of the soft branches' linear excess
risk therefore concerns access to retained predictive information.

\begin{samepage}
\paragraph{Task dependence.}
When \(U\) is irrelevant to the token (\(b=0\)), the
\(n=512,\sigma=0.01\) linear JEPA-minus-NTP difference is
\(-0.016546\), reversing the ranking. This diagnostic uses one encoder
seed and two repeats.
\par\end{samepage}

\begin{samepage}
These comparisons distinguish information retained by an encoder,
information available through the observation channel, and information
accessible to a fitted readout.
\par\end{samepage}

\begin{samepage}
\section{Implications for JEPA objectives and evaluation}
\label{sec:implications}
Global content selection can be certified by comparing a lower bound over
the entire invariant feasible class with a feasible encoder retaining
observation variation. The scalar model uses an atom-count comparison;
under complete scene rank and whitening, the scene Gram matrix reduces the
invariant class to one rotation orbit.
\par\end{samepage}

Whitening fixes covariance; Gaussian EP rewards marginal shape beyond
second moments. In the high-dimensional model, scene probabilities constrain
that shape, and independent nuisance improves it under the same whitening
constraint. Thus the regularizer on the invariant class and the view
transition's spectrum both enter the analysis.

For evaluation, scene alignment measures agreement with a selected latent
variable; observation Bayes risk measures task information available through
the specified channel; the fitted head gap measures how well a readout uses
it. Their joint reporting separates geometric agreement, stable task
information, and learnability from finite labels.

\begin{samepage}
\section{Conclusion}
\label{sec:conclusion}
Gaussian marginal matching participates directly in representation content
selection. Our global constructions show that it can force independently
resampled variation into optimal representations, including under exact
whitening and sufficient scene spectral rank. The scalar model also yields
equal-value global minimizers outside a single sign orbit. The readout
study shows why the value of retained variation depends on the prediction
task and the observation channel. The joint objective selects representation
content; the prediction task and observation channel determine its usable value.
\par\end{samepage}

\clearpage
\setlength{\textfloatsep}{20pt plus 2pt minus 4pt}
\setlength{\intextsep}{12pt plus 2pt minus 2pt}
\setlength{\abovecaptionskip}{10pt}
\begingroup
\small
\setlength{\bibsep}{0pt}

\endgroup
\clearpage
\appendix
\section{Objectives and normalization}
\label{app:objectives}
For a finite observation law \(p\), a symmetric positive-pair law \(\pi\)
with marginals \(p\), and an encoder \(h\), define the pair energy
\[
 \mathcal E_\pi(h)=\sum_{i,j}\pi_{ij}\norm{h_i-h_j}^2.
\]
With two global views only, the mean squared distance to their mean is
\(P=\mathcal E_\pi/4\) for scalar outputs. For \(V\) same-scene views and
\(G\) global views, our coordinate-mean prediction term is
\begin{equation}
 P_{G,V}=\E\frac1{Vk}\sum_{v=1}^V
 \norm{h_v-\bar h_G}^2,\qquad
 \bar h_G=\frac1G\sum_{g=1}^G h_g.
 \label{eq:center}
\end{equation}
The general-view coefficient will be derived explicitly below.

For a scalar law \(\mu\) write \(\cf_\mu(t)=\E_\mu e^{itX}\) and
\(g(t)=e^{-t^2/2}\). For a positive frequency measure \(\nu\), set
\begin{equation}
 D_\nu(\mu)=\int|\cf_\mu(t)-g(t)|^2\,d\nu(t),\quad
 A_\nu=\int d\nu(t),\quad
 J_\nu(\mu)=\int|\cf_\mu(t)|^2\,d\nu(t).
 \label{eq:disc}
\end{equation}
For an iid batch of size \(N\), the EP statistic is \(N\) times the
empirical discrepancy. Its exact expectation is
\begin{equation}
 R_{N,\nu}(\mu)=ND_\nu(\mu)+A_\nu-J_\nu(\mu).
 \label{eq:iid}
\end{equation}
Indeed,
\(\E|\widehat\cf_N(t)|^2=N^{-1}+(1-N^{-1})|\cf_\mu(t)|^2\);
the empirical target cross term is unbiased. Thus the discrepancy bias is
\((A_\nu-J_\nu)/N\), whereas the scaled EP bias is \(A_\nu-J_\nu\).

\begin{tablehere}
\centering\small
\begin{tabular}{>{\raggedright\arraybackslash}p{0.19\linewidth}>{\raggedright\arraybackslash}p{0.33\linewidth}>{\raggedright\arraybackslash}p{0.38\linewidth}}
\toprule
Object & Frequency measure & Sampling and objective\\
\midrule
Scalar analytic &
\(d\nu=e^{-t^2/2}dt,\ t\in\R\) &
\(P+\gamma D_\nu\), or \(P+\alpha R_{N,\nu}\);
two views, scalar outputs.\\[3pt]
Scalar quadrature &
\(t_j=5j/8,\ -8\leq j\leq8\);
trapezoidal weights times \(g_j\) &
Per-view iid size \(N\); EP averaged over two views.
Mixed weight \(\lambda\) gives \(\alpha=\lambda/(1-\lambda)\).\\[3pt]
Pooled quadrature &
\(t_j=3j/16,\ 0\leq j\leq16\);
symmetry-doubled trapezoidal weights times \(g_j\) &
Eight same-scene views, two global views, \(N=256\),
\(M=NV=2048\); \(P+\lambda R_{\rm pooled}\), \(\lambda=1/50\).\\
\bottomrule
\end{tabular}
\caption{Three different objectives. The pooled convention follows the
\texttt{stable-pretraining} 0.1.8 release \citep{stable}. Its weight \(1/50\)
is a direct coefficient and is not the scalar mixed-loss coefficient
after division by \(1-\lambda\).}
\label{tab:objectives}
\end{tablehere}

For vector outputs, the published-grid discrepancy is averaged over
\(\theta\sim\mathrm{Unif}(\mathbb S^{k-1})\) after projection onto
\(\theta^\top h\). The high-dimensional theorems use the \emph{exact}
direction expectation. They do not use a finite set of sampled directions.
Exact population whitening means \(\E h=0\), \(\E hh^\top=I_k\).
It is a constraint on a fixed observation encoder, distinct from
training-mode BatchNorm.

The alignment metric for two encoders is
\(\min_{Q\in O(k)}\E\norm{h_1-Qh_2}^2/k\), with independent modality
nuisances conditional on the same scene. Scalar orbit distance without
rescaling is
\(d_\pm(x,z)=\min\{\norm{x-z}_p,\norm{x+z}_p\}\).
The readout experiment separately measures clean alignment to a chosen
scene variable and noisy observation risk. These metrics are not
interchangeable.

\section{Complete scalar proof and variants}
\label{app:scalar}
\subsection{Pair law and the role of symmetry}
Index four observations by
\((S,U)=(1,1),(1,-1),(-1,1),(-1,-1)\), with uniform \(p_i=1/4\).
Let \(s=(1,1,-1,-1)\), \(\eta=1/2\), and
\begin{equation}
 \pi_{ij}=\frac{1+\eta s_i s_j}{16},\qquad K_{ij}=4\pi_{ij}.
 \label{eq:pair}
\end{equation}
All pair probabilities are strictly positive. The observation transition
has eigenvalues \(1,\eta,0,0\). Equivalently, draw correlated scenes with
\(\Pr(S=s,S'=s')=(1+\eta ss')/4\), and resample the two nuisance signs
independently. Any injective modality rendering gives the same world law.
Thus the nuisance has no positive-pair correlation.

For \(x\in\R^4\), let \(m=\E_p x\), \(v=\Var_p x\), and
\(a=\E_p Sx\). Expansion gives
\begin{equation}
 \mathcal E_\pi(x)=2v-2\eta a^2,\qquad
 P_\pi(x)=v/2-a^2/4\geq v/4.
 \label{eq:scalarP}
\end{equation}
The within-group swaps \(T_+=(12)\), \(T_-=(34)\) preserve \(p\)
and \(\pi\). They therefore preserve both \(P_\pi\) and every regularizer
depending only on \(\mu_x=\frac14\sum_i\delta_{x_i}\).
This follows by a change of indices in the pair sum and the pushforward
measure. Scalar rotations form only the orbit \(\{x,-x\}\).
The next lemma forces the swaps to act outside that orbit at a minimum.

\subsection{Separation between two and four atoms}
For the analytic measure \(d\nu=e^{-t^2/2}dt\), put
\(A=\sqrt{2\pi}\), \(B=\sqrt\pi\), \(C=\sqrt{2\pi/3}\).
Gaussian integration gives
\begin{equation}
 D_{\rm an}(\mu_x)=\frac A{16}\sum_{i,j}e^{-(x_i-x_j)^2/2}
       -\frac B2\sum_i e^{-x_i^2/4}+C.
 \label{eq:analyticD}
\end{equation}
For the listing grid define
\begin{equation}
 t_j=5j/8,\quad
 \omega_j=(5/8)c_j e^{-t_j^2/2},\quad -8\leq j\leq8,
 \qquad c_{\pm8}=1/2,\quad c_j=1\ (|j|<8).
 \label{eq:scalargrid}
\end{equation}
Then \(D_{\rm gr}=\sum_j\omega_j|\cf(t_j)-g_j|^2\).

\begin{lemma}[Global regularizer separation]
\label{lem:separation}
For either \(D=D_{\rm an}\) or \(D=D_{\rm gr}\), every equally weighted
two-atom law satisfies \(D((\delta_b+\delta_c)/2)>1/50\).
The rational vector
\[
 w=(1/10,13/10,-1/10,-13/10)
\]
satisfies \(P_\pi(w)=121/400\), \(D(\mu_w)<3/1000\), and
\(A_\nu<251/100\).
For the analytic discrepancy its exact two-atom minimum is
\begin{equation}
 d_*=\frac{\sqrt{2\pi}}2+\sqrt{\frac{2\pi}3}
       -\frac74\sqrt\pi\,2^{-3/14}
       >\frac{357}{16000}>\frac1{50}.
 \label{eq:dstar}
\end{equation}
\end{lemma}
\begin{proof}
For two atoms \(b=q+r,c=q-r\), \(r\geq0\), the analytic discrepancy is
\[
 \frac A2(1+e^{-2r^2})-B\,U(q,r)+C,\qquad
 U(q,r)=2e^{-(q^2+r^2)/4}\cosh(qr/2).
\]
The bound \(\cosh z\leq e^{z^2/2}\) follows by integrating
\(\tanh z\leq z\) on \(z\geq0\). If \(r^2\leq2\), then
\(U(q,r)\leq2e^{-r^2/4}\). Equality is attained at \(q=0\).
The resulting lower function
\[
 f(r)=\frac A2(1+e^{-2r^2})-2B e^{-r^2/4}+C
\]
has derivative
\(f'(r)=r e^{-r^2/4}(B-2Ae^{-7r^2/4})\).
It decreases to \(r_*^2=(6/7)\log2<2\), then increases.
Substitution gives \eqref{eq:dstar}.

If \(r\geq\sqrt2\), scale both atoms by \(c_0=\sqrt2/r\leq1\).
This increases each target exponential and makes the half-separation
\(\sqrt2\), so \(U(q,r)\leq2e^{-1/2}\).
The discrepancy is therefore at least \(A/2+C-2Be^{-1/2}>d_*\).
To check the strict inequality, \(z=2^{-3/14}>4/5\) because
\((4/5)^{14}<1/8\), and \(e^{-1/2}<2/3\), so
\((7/4)z-2e^{-1/2}>7/5-4/3>0\).
Also \(1/\sqrt2>707/1000\), \(\sqrt{2/3}>816/1000\),
\(z<863/1000\), and \(\sqrt\pi>7/4\). Thus
\[
 d_*>\frac74\left(\frac{1523}{1000}
                  -\frac74\frac{863}{1000}\right)=\frac{357}{16000}.
\]
All auxiliary comparisons reduce to integer powers or squares; the
exponential comparison follows from \(e^{1/2}>1+1/2\).

For the grid, each atom is periodic with period \(16\pi/5\).
Writing \(b=8\pi u/5,c=8\pi v/5\), it suffices to cover
\([-1,1]^2\). Partition that square into 154 rational
rectangles. On each rectangle, outward enclosures of the cosine and sine
at all 17 signed nodes give a discrepancy lower bound strictly exceeding
\(1/50\). Exact rational sweeps check complete coverage without interior
gaps or overlaps. Appendix~\ref{app:cert} states the interval algorithm.
The smallest certified rectangle bound exceeds \(0.0200111\);
the complete rational partition and certified lower bounds are listed in
Appendix~\ref{app:partition}, Table~\ref{tab:scalar_partition}.

The prediction value of \(w\) is exact arithmetic. Outward evaluation of
\eqref{eq:analyticD} and \eqref{eq:scalargrid} gives respectively
\[
 0.00216433<D_{\rm an}(\mu_w)<0.00216434,\qquad
 0.00216238<D_{\rm gr}(\mu_w)<0.00216239.
\]
Both are below \(3/1000\); the grid window mass is below \(2.506626\).
\end{proof}

\subsection{Attainment and nonorthogonal global minima}
\begin{theorem}[Scalar soft-objective nonuniqueness]
\label{thm:scalar}
On \(\R^4\), either objective \(P_\pi+\gamma D_{\rm an}\) or
\(P_\pi+\gamma D_{\rm gr}\) attains a global minimum and has, for every
global minimizer \(x\), an equal-value within-group swap outside
\(\{x,-x\}\) whenever \(\gamma\geq20\). The analytic conclusion also holds
for \(\gamma\geq16\).

For either frequency measure, the same conclusion holds for
\(P_\pi+\alpha R_{N,\nu}\) if
\begin{equation}
 \alpha\left(\frac{17N}{1000}-\frac{251}{100}\right)>\frac{121}{400}.
 \label{eq:weight}
\end{equation}
All these statements remain valid on
\(X_0=\{x:\E_p x=0\}\).
\end{theorem}
\begin{proof}
A group-constant vector \((b,b,c,c)\) has a two-atom law.
By Lemma~\ref{lem:separation}, its deterministic objective is greater
than \(\gamma/50\), while
\[
 F_\gamma(w)<121/400+3\gamma/1000<\gamma/50
 \quad(\gamma\geq20).
\]
For the analytic case use the larger lower bound \(\gamma d_*\);
\eqref{eq:dstar} shows the witness is below it for \(\gamma\geq16\).
For the iid statistic, \(0\leq J_\nu\leq A_\nu\) and
\eqref{eq:iid} imply a group-constant lower bound \(\alpha N/50\).
The witness upper bound is
\[
 121/400+\alpha(3N/1000+251/100)<\alpha N/50,
\]
by \eqref{eq:weight}. Thus every minimum, if attained, splits a group.

On \(X_0\), nonnegative regularization and \eqref{eq:scalarP} make each
objective coercive. For the full domain, bounded objective gives bounded
variance and hence bounded centered coordinates. With the grid, translate
all coordinates by integer multiples of \(16\pi/5\) until the mean lies in
\([-8\pi/5,8\pi/5]\). This preserves prediction and every node CF, making
a minimizing sequence bounded.

For the analytic deterministic objective, an unbounded mean with bounded
centered coordinates makes the target cross term in
\eqref{eq:analyticD} vanish. The escape lower bound is \(\gamma C\),
above the witness threshold. In the iid case, writing
\(X_\nu=B\E e^{-X^2/4}\) for the analytic target cross integral gives
\[
 R_{N,\nu}=(N-1)J_\nu-2NX_\nu+NC+A.
\]
Condition~\eqref{eq:weight} implies \(N>1\). Mean escape has
lower bound \(NC+A\), again above the witness threshold \(\alpha N/50\).
Every case therefore has a bounded minimizing sequence. Continuity
proves attainment.

A nonconstant optimizer has \(a=\E Sx\neq0\): otherwise assign the two
largest values to \(S=1\) and the other two to \(S=-1\). This preserves the
law and variance, gives nonzero correlation, and strictly lowers
\eqref{eq:scalarP}. Choose a split group and exchange its unequal values.
The swap preserves the objective and differs from \(x\). If it equalled
\(-x\), the untouched values would be zero and the exchanged values
opposites, forcing \(a=0\). It therefore lies outside the sign orbit.
Every permutation used here preserves \(X_0\).
\end{proof}

For a mixed two-view loss \((1-\lambda)P+\lambda R_N\),
\(\alpha=\lambda/(1-\lambda)\). Condition~\eqref{eq:weight} includes
\((N,\lambda)=(256,1/5)\) and \((1024,1/50)\). For a deterministic ECF
over all four states, \(\gamma=4\lambda/(1-\lambda)\); the sufficient
mixed weights are \(5/6\) for the grid and \(4/5\) for the analytic
improvement. These are sufficient conditions, not exact phase transitions.

\subsection{Scalar quadrature and global phase coverage}
\label{app:cert}
Intervals are represented by integer endpoints at scale \(10^{90}\).
Addition, multiplication, and division round outward, and square roots
are enclosed using integer square roots. To enclose trigonometric
arguments, use Machin's formula
\[
 \pi=16\arctan(1/5)-4\arctan(1/239).
\]
Eighty terms of each alternating arctangent series, with its next-term
remainder, enclose \(\pi\). For \(\cos(\pi q)\), exact rational
periodic reduction puts the argument in \([0,\pi/2]\); a 160-term
cosine series encloses endpoint values. Each integer-\(q\) extremum
inside an interval is included. Sine is evaluated as
\(\cos(\pi(q-1/2))\). Gaussian weights use a 160-term exponential
series with a next-term error bound; large exponents are reduced by
halving and the result squared with outward rounding.

For a rectangle \(I_u\times I_v\), at each signed node form interval
enclosures of
\[
 \frac{\cos(j\pi u)+\cos(j\pi v)}2-g_j,\qquad
 \frac{\sin(j\pi u)+\sin(j\pi v)}2.
\]
If \(I=[a,b]\), a lower bound on a squared term is
\[
 \ell(I)=
 \begin{cases}
 0,&a\leq0\leq b,\\
 \min(a^2,b^2),&\text{otherwise}.
 \end{cases}
\]
Summing the weighted lower bounds gives a discrepancy lower bound
for every pair of atoms in the rectangle. A partition of
\([-1,1]^2\) into 154 rational rectangles has minimum bound greater
than \(0.0200111>1/50\).
Complete coverage is verified by sweeping the rational vertical
endpoints: in each slab the active vertical intervals concatenate
from \(-1\) to \(1\), with disjoint interiors.
Periodicity of each atom then covers all of \(\R^2\).
The exact rectangle endpoints and their outward lower bounds are
listed in Appendix~\ref{app:partition}, Table~\ref{tab:scalar_partition}.

The same arithmetic evaluates the four-atom witness and frequency
moments:
\[
\begin{aligned}
 0.00216433&<D_{\rm an}(\mu_w)<0.00216434,\\
 0.00216238&<D_{\rm gr}(\mu_w)<0.00216239,\\
 A_{\rm gr}&<2.506626,\qquad
 M_2<2.51,\qquad M_{2,g}<0.89.
\end{aligned}
\]

\section{Quantitative stability}
\label{app:stability}

For \(x\in X_0\), put \(u=\E\Var(x\mid S)=v-a^2\) and \(y=\E[x\mid S]=aS\).
Then \(P_\pi(x)=a^2/4+u/2\) and \(P_\pi(x)-P_\pi(y)=u/2\).
Theorem~\ref{thm:scalar} already proves \(u>0\) at all its optima.
The following constants also quantify separation for one grid objective.

\begin{lemma}[Projection and orbit bounds]
\label{lem:projection}
Fix \(F_\pi=P_\pi+R_{1024,\rm gr}/49\) on \(X_0\).
Define
\[
 H_0=\frac{184273}{500000},\quad B_0=\frac{1024}{2450},
 \quad L_0=\frac{345459}{4900}.
\]
Then \(F_\pi(w)<H_0\) and, for every \(x\in X_0\),
\begin{equation}
 B_0<F_\pi(\E[x\mid S])\leq F_\pi(x)+L_0u(x).
 \label{eq:projection}
\end{equation}
If \(m_*=\min_{X_0}F_\pi\) and \(F_\pi(x)-m_*\leq\delta\), then
\begin{equation}
 \max_{T\in\{T_+,T_-\}}d_\pm(x,Tx)^2\geq2u(x)-24\delta.
 \label{eq:orbit}
\end{equation}
Every base minimizer has \(u>7/10000\).
\end{lemma}
\begin{proof}
Write
\(x=(a+d_+,a-d_+,-a+d_-,-a-d_-)\);
\(u=(d_+^2+d_-^2)/2\).
The conditional cosine factors and \(1-\cos z\leq z^2/2\) give
\(|\cf_x(t)-\cf_y(t)|\leq t^2u/2\).
The certified moments
\[
 M_2=\sum_j\omega_jt_j^2<251/100,\qquad
 M_{2,g}=\sum_j\omega_jg_jt_j^2<89/100
\]
bound the self-term change by \(M_2u\) and target cross-term change by
\(M_{2,g}u/2\). Expanding \eqref{eq:iid},
\[
 \frac1{49}|R_{1024,\rm gr}(x)-R_{1024,\rm gr}(y)|
 \leq\frac{1023(251/100)+1024(89/100)}{49}u
 =\frac{347909}{4900}u.
\]
Subtracting the prediction decrease \(u/2\) proves the upper comparison
in \eqref{eq:projection}; Lemma~\ref{lem:separation} proves the lower one.
At a minimizer \(F_\pi(x)<H_0\), this gives the stated rational bound on
\(u\).

For the orbit bound, the three balanced vectors
\((1,1,-1,-1),(1,-1,1,-1),(1,-1,-1,1)\) are an orthonormal basis of \(X_0\).
Their correlations obey \(\sum_i a_i^2=v\).
A permutation can therefore give scene correlation
\(a_*^2=\max_i a_i^2\geq v/3\).
It preserves the law and variance, and lowers the objective by
\((a_*^2-a^2)/4\leq\delta\). Consequently \(4a^2\geq2u-24\delta\).
Directly,
\[
 \begin{split}
 d_\pm(x,T_+x)^2&=\min(2d_+^2,4a^2+2d_-^2),\\
 d_\pm(x,T_-x)^2&=\min(2d_-^2,4a^2+2d_+^2).
 \end{split}
\]
Choose the larger \(|d_\pm|\); the maximum is at least
\(\min(2u,4a^2)\geq2u-24\delta\).
\end{proof}

\begin{corollary}[Explicit separated partners after perturbation]
\label{cor:perturb}
For the objective in Lemma~\ref{lem:projection}, let \(\pi'\) be a symmetric
pair law with uniform marginals and
\(\eps=\sum_{i,j}|\pi'_{ij}-\pi_{ij}|\leq1/200000\).
Every global minimizer \(y\) of \(F_{\pi'}\) on \(X_0\) has a within-group
swap \(z=Ty\) satisfying
\[
 d_\pm(y,z)>1/40,\qquad
 0\leq F_{\pi'}(z)-\min F_{\pi'}<3/100000.
\]
More precisely the excess is at most \(16\eps\tau H_0\), where
\(\tau=(1+8\eps)/(1-8\eps)\).
\end{corollary}
\begin{proof}
Since \(\operatorname{diam}(x)^2\leq8\norm{x}_p^2\) and
\(P_\pi\geq\norm{x}_p^2/4\),
\[
 |P_{\pi'}-P_\pi|\leq2\eps\norm{x}_p^2\leq8\eps P_\pi.
\]
The regularizer is unchanged and nonnegative, so
\((1-8\eps)F_\pi\leq F_{\pi'}\leq(1+8\eps)F_\pi\).
This proves coercivity and attainment. For every perturbed minimizer,
\[
 F_\pi(y)<\tau H_0,\quad
 F_\pi(y)-m_*\leq\Delta_\eps:=\frac{16\eps H_0}{1-8\eps},
 \quad u(y)>\frac{B_0-\tau H_0}{L_0}=:u_\eps.
\]
Lemma~\ref{lem:projection} gives a squared distance greater than
\(2u_\eps-24\Delta_\eps\). This expression decreases with \(\eps\), and
the constants in Lemma~\ref{lem:projection} give
\[
 2u_{1/200000}-24\Delta_{1/200000}>\frac1{1600}.
\]
Each original swap preserves \(F_\pi\). Its perturbed excess is at most
\(4\eps\norm{y}_p^2\leq16\eps\tau H_0\).
At \(\eps=1/200000\), this upper bound is less than \(3/100000\).
\end{proof}

At \(\eps=0\) the partner is another exact optimum; at positive
\(\eps\) only near-optimality is asserted. The perturbation preserves a
spectral gap: \(\norm{4(\pi'-\pi)}_{\rm op}\leq4\eps\), so the gap on the
centered space is at least \(1/2-8\eps\).
It need not preserve the full independent-nuisance good-view structure.
For example \(\pi'=\pi+(\eps/64)bb^\top\), \(b=(3,1,-1,-3)\), has uniform
marginals, total entrywise difference \(\eps\), and breaks both swaps.

The scalar base optima also have a normalized
alignment consequence. Since \(v<4H_0\),
\[
 u/v>c_0:=\frac{B_0-H_0}{4H_0L_0}>\frac{19}{40000}.
\]
With independent modality nuisances on the same scene, two variance-one
rescaled optima have best-sign MSE
\(2-2|a_1a_2|/\sqrt{v_1v_2}>2c_0>19/20000\).
This is evaluation normalization, not an added whitening constraint on
the soft optimization problem.

\section{Complete high-dimensional proof}
\label{app:highdim}
\subsection{A finite scene model with sufficient spectral rank}
Let \(k=512\) and let \(S\) have 513 possible values with
\begin{equation}
 p_1=p_2=1/4,\qquad p_s=1/1022\quad(3\leq s\leq513).
 \label{eq:sceneprobs}
\end{equation}
The remaining 511 scenes jointly carry half the population. Let
\(U\sim\mathrm{Unif}\{-1,1\}^k\), independent of \(S\), and let each
positive view draw a fresh independent \(U\) while keeping the same scene.
An injective rendering is \(g(S,U)=(r(S),U)\), where \(r\) is any complete
orthonormal basis of the centered \(L^2(p)\) scene functions.
The positive-view transition on observations is conditional expectation
given \(S\). It has 512 nonconstant scene eigenvalues equal to one and
all remaining nuisance eigenvalues zero. There is a gap after the 512th
nonconstant direction.

Optimize over all finite-support encoders \(h(S,U)\in\R^k\) satisfying
\begin{equation}
 \E h=0,\qquad \E hh^\top=I_k,\qquad
 u(h)=\frac1k\E\norm{h-\E[h\mid S]}^2.
 \label{eq:whiten}
\end{equation}
The quantity \(u\in[0,1]\) is the fraction of coordinate-mean variance
due to independent nuisance. For conditionally independent views,
\eqref{eq:center} has expectation
\begin{equation}
 P_{G,V}=\left(1+\frac1G-\frac2V\right)u.
 \label{eq:viewcoef}
\end{equation}
For a global view its residual variance is \((1-1/G)ku\); for a
nonglobal view it is \((1+1/G)ku\). Averaging gives
\eqref{eq:viewcoef}. We fix \(G=2,V=8,N=256\); thus \(P=5u/4\).

\subsection{The pooled statistic and its expectation}
The \texttt{stable-pretraining} 0.1.8 release uses a merged view
axis for EP \citep{stable}. Set \(M=NV=2048\) and
\begin{equation}
 t_j=3j/16,\quad
 \omega_j=(3/16)d_jg_j,\quad
 g_j=e^{-t_j^2/2},\quad
 d_0=d_{16}=1,\quad d_j=2\ (1\leq j\leq15).
 \label{eq:poolednodes}
\end{equation}
For the merged batch, let
\[
 \widehat R_{\mathcal B,\theta}(h)=
 M\sum_{j=0}^{16}\omega_j
 \left|\frac1M\sum_{i=1}^N\sum_{v=1}^V
 e^{it_j\theta^\top h(S_i,U_{iv})}-g_j\right|^2.
\]
The weights combine the two signs of a symmetric trapezoidal grid.
Directions are uniform on \(\mathbb S^{k-1}\).

Write the direction-averaged marginal and conditional self terms as
\[
 J(h)=\sum_j\omega_j\E_\theta|\cf_{h,\theta}(t_j)|^2,\qquad
 J_c(h)=\sum_j\omega_j\E_{S,\theta}
                |\cf_{h\mid S,\theta}(t_j)|^2.
\]
Let \(A_\nu=\sum_j\omega_j\),
\(X(h)=\sum_j\omega_jg_j\E_\theta\operatorname{Re}\cf_{h,\theta}(t_j)\),
\(C_\nu=\sum_j\omega_jg_j^2\), and \(D=J-2X+C_\nu\).
Independent examples have product CF \(J\), while two distinct views
of the same example have the conditional term \(J_c\).
Counting \(M\) diagonal terms and \(NV(V-1)\) within-example off-diagonal
terms gives the exact expectation
\begin{align}
 \bar R(h)&:=\E_{\mathcal B,\theta}\widehat R_{\mathcal B,\theta}(h)
       =MD+(A_\nu-J)+(V-1)(J_c-J)\notag\\
       &=(M-V)J+(V-1)J_c-2MX+MC_\nu+A_\nu.
 \label{eq:pooledexpect}
\end{align}
The within-example term cannot be replaced by an iid batch of \(M\)
unrelated observations. The population objective studied here is
\begin{equation}
 F(h)=\frac54u(h)+\frac1{50}\bar R(h).
 \label{eq:Fhigh}
\end{equation}
This follows the release's direct \(P+\lambda EP\) coefficient convention,
with an \emph{additional} exact population-whitening constraint.

\subsection{A gap covering the whole scene-only class}
\begin{lemma}[Complete scene Gram matrix]
\label{lem:gram}
Every scene-only encoder satisfying \eqref{eq:whiten} has the form \(Qr(S)\)
for some \(Q\in O(k)\), and
\[
 r(s)^\top r(s')=\frac{\mathbf1_{s=s'}}{p_s}-1.
\]
Its squared norms are 3 on the two common scenes and 1021 on the other
scenes. Distinct-scene squared distances are 8, 1026, and 2044 for the
common/common, common/other, and other/other cases.
\end{lemma}
\begin{proof}
The \(513\times512\) matrix with rows \(\sqrt{p_s}r(s)^\top\) has
orthonormal columns spanning \(\sqrt p^\perp\). Its projection is
\(I-\sqrt p\sqrt p^\top\), proving the Gram identity.
Every centered scene function has coefficients in this complete basis.
Its covariance is identity precisely when its coefficient matrix is
orthogonal. Norms and distances follow by substitution.
\end{proof}

The sphere characteristic function of a vector with squared norm \(d\) is
\begin{equation}
 \psi_d(t):=\E_\theta e^{it\theta^\top a}
 ={}_0F_1(;256;-t^2d/4),\qquad \norm a^2=d.
 \label{eq:spherecf}
\end{equation}
Expanding the exponential and using the sphere's even coordinate moments
gives this series; its value depends only on \(d\).
Lemma~\ref{lem:gram} therefore makes the entire scene-only class share
the following self and cross factors:
\begin{align}
 j_s(t)&=\frac18+\frac1{2044}+\frac18\psi_8(t)
      +\frac{255}{1022}\psi_{2044}(t)+\frac12\psi_{1026}(t),\notag\\
 x_s(t)&=\tfrac12\psi_3(t)+\tfrac12\psi_{1021}(t).
 \label{eq:scenefactors}
\end{align}
The constant terms count identical independent scenes; the other
coefficients count the ordered distinct-scene pairs.

Use the feasible nuisance witness \(h_U=U\).
For two independent cube points the Hamming distance is
\(\mathrm{Bin}(512,1/2)\), so
\begin{equation}
 j_U(t)=2^{-512}\sum_{\ell=0}^{512}\binom{512}{\ell}\psi_{4\ell}(t),
 \qquad x_U(t)=\psi_{512}(t).
 \label{eq:cubefactors}
\end{equation}
For scene-only encoders \(J_c=A_\nu\); for the scene-independent cube
\(J_c=J_U\). Consequently their objectives are
\begin{equation}
 F_s=\frac1{50}[MD_s+V(A_\nu-J_s)],\qquad
 F_U=\frac54+\frac1{50}[MD_U+A_\nu-J_U].
 \label{eq:comparators}
\end{equation}

\begin{table}[tb]
\centering
\begin{tabular}{lrr}
\toprule
Quantity & All scene-only whitened encoders & Cube witness\\
\midrule
Direction-averaged discrepancy \(D\) & 0.0497120941 & 0.000000450006\\
Expected pooled EP & 108.79885252 & 1.05385614\\
Expected prediction & 0 & 1.25\\
Objective & 2.17597705 & 1.27107712\\
\bottomrule
\end{tabular}
\caption{Objective values for the scene-only class and cube comparator,
rounded from interval enclosures. The strict bounds are
\(F_s>2.1759\), \(F_U<1.2711\), and \(F_s-F_U>9/10\).}
\label{tab:gap}
\end{table}

Appendix~\ref{app:sphere} specifies an outward series evaluation of
\eqref{eq:spherecf} covering every node and squared distance. It certifies
the strict bounds in Table~\ref{tab:gap}. Thus no scene-only encoder
is a global minimizer. The next theorem quantifies how far every optimum
must depart from that class.

\subsection{A uniform lower bound for every population optimum}
\begin{theorem}[Whitening plus pooled EP forces nuisance]
\label{thm:high}
The objective \eqref{eq:Fhigh} attains its minimum over
\eqref{eq:whiten}. Every global minimizer satisfies \(u(h)>3/20000\).
For any two global minimizers evaluated on the same scene with independent
modality nuisances,
\[
 \min_{Q\in O(k)}\E\norm{h_1-Qh_2}^2/k>3/10000.
\]
The same statements hold in the linear class
\(h_W=W(r,U)\), \(W\in\R^{512\times1024}\), \(WW^\top=I_{512}\).
\end{theorem}
\begin{proof}
Write \(y=\E[h\mid S]=Ar(S)\), \(z=h-y\), and \(C_n=\E zz^\top\).
Whitening gives \(AA^\top=I-C_n\), \(0\preceq C_n\preceq I\),
and \(\tr C_n=ku\). Extend a polar factor of \(A\) to an orthogonal
matrix \(Q\), also when \(A\) is singular. If \(c_i\) are the eigenvalues
of \(C_n\), then
\begin{equation}
 \frac1k\E\norm{y-Qr}^2
 =\frac1k\sum_i(1-\sqrt{1-c_i})^2
 \leq\frac1k\sum_i c_i^2\leq ku^2.
 \label{eq:polar}
\end{equation}
The first-order remainder from \(z\) cancels conditional on \(S\).
Using \(|e^{iv}-1-iv|\leq v^2/2\) and
\(\E_\theta\theta\theta^\top=I/k\),
\[
 \E_\theta|\cf_h(t)-\cf_y(t)|\leq t^2u/2.
\]
Cauchy--Schwarz and \eqref{eq:polar} give
\(\E_\theta|\cf_y(t)-\cf_{Qr}(t)|\leq |t|\sqrt{k}\,u\).
The combined bound is \(b_tu\), where \(b_t=t^2/2+|t|\sqrt{k}\).
Because CF magnitudes are at most one,
\[
 |J(h)-J_s|\leq2\sum_j\omega_jb_{t_j}u,\qquad
 |X(h)-X_s|\leq\sum_j\omega_jg_jb_{t_j}u.
\]
For the conditional self term take two independent copies of \(z\)
given \(S\) and apply \(1-\cos v\leq v^2/2\). Direction averaging yields
\(0\leq A_\nu-J_c(h)\leq M_2u\), \(M_2=\sum_j\omega_jt_j^2\).

Define \(M_1=\sum_j\omega_j|t_j|\),
\(M_{1,g}=\sum_j\omega_jg_j|t_j|\), and
\(M_{2,g}=\sum_j\omega_jg_jt_j^2\).
Equation~\eqref{eq:pooledexpect} now gives
\begin{equation}
 |\bar R(h)-\bar R_s|\leq K_0u,\quad
 K_0=(M-V)(M_2+2\sqrt{k}M_1)
      +M(M_{2,g}+2\sqrt{k}M_{1,g})+(V-1)M_2.
 \label{eq:K}
\end{equation}
Outward evaluation of these finite sums gives
\(0<K_0/50-5/4<6000\).

The observation support is finite and each atom has positive mass.
The whitening constraint bounds every output coordinate at every atom,
and defines a nonempty closed set. Hence the feasible domain is compact.
The objective is continuous, so a minimum exists. The cube is feasible,
and every minimum has \(F(h)\leq F_U\). Equations~\eqref{eq:Fhigh}
and~\eqref{eq:K} imply
\[
 F_s\leq F(h)+(K_0/50-5/4)u.
\]
Together with \(F_s-F_U>9/10\), this gives
\(u>(9/10)/6000=3/20000\).

Conditional independence eliminates nuisance from the cross-modal inner
product. Since \(\E\norm{y_i}^2=k(1-u_i)\), for every orthogonal \(Q\)
Cauchy--Schwarz gives
\[
 \E\langle h_1,Qh_2\rangle/k
 =\E\langle y_1,Qy_2\rangle/k
 \leq\sqrt{(1-u_1)(1-u_2)}.
\]
Both total variances equal \(k\), so the coordinate-mean MSE is at least
\(2-2\sqrt{(1-u_1)(1-u_2)}>3/10000\).

Finally, \((r,U)\) is centered with covariance \(I_{1024}\) and is
injective. The constrained linear parameter domain is compact and
contains all scene-only witnesses \([Q,0]\) and the cube \([0,I]\).
The same global comparisons apply in that class.
\end{proof}

\subsection{Sphere characteristic function and tail bounds}
\label{app:sphere}

For \(\theta\) uniform on \(\mathbb S^{k-1}\),
\(\E\theta_1^{2m}=(1/2)_m/(k/2)_m\), and odd moments vanish.
Consequently the characteristic function of a vector of squared
norm \(d\) is
\[
 \psi_d(t)=\sum_{m=0}^{\infty}
 \frac{(-a)^m}{m!(k/2)_m},\qquad a=t^2d/4.
\]
All high-dimensional evaluations have \(k=512\) and
\(0\leq a\leq4608\). After 128 nonconstant terms,
\[
 |\mathcal R_{128}|
 \leq\frac{a^{129}}{129!(256)_{129}(1-q)},\qquad
 q=\frac{4608}{130\cdot385}<\frac1{10}.
\]
This remainder is uniformly below \(1.44\times10^{-68}\).
An evaluation with 160 nonconstant terms uses first omitted term
161 and ratio bound \(4608/(162\cdot417)\).
Each finite sum is enclosed with outward rounding and enlarged
by its symmetric tail bound. Cube weights are exact binomial
fractions. Direction averaging is therefore evaluated analytically
through \(\psi_d\).

\begin{table}[ht]
\centering\small
\begin{tabular}{lr}
\toprule
Quantity & Outward enclosure\\
\midrule
Population gap \(F_s-F_U\) & \((0.90489992,\ 0.90489994)\)\\
Population coefficient \(K_0/50-5/4\) & \((5617,\ 5618)\)\\
Batch quadratic \(\alpha_0\) & \((0.72315443,\ 0.72315444)\)\\
Batch quadratic \(\beta_0\) & \((-0.38850033,\ -0.38850032)\)\\
Batch quadratic \(\eta_0\) & \((0.06317364,\ 0.06317366)\)\\
Scene lower bound at \(117/256\) & \((1.50190470,\ 1.50190471)\)\\
Scene-count failure probability & \((0.07521726,\ 0.07521727)\)\\
Cube Hoeffding failure bound & \((0.00000871169,\ 0.00000871170)\)\\
Cube Markov failure bound & \((0.09580510,\ 0.09580511)\)\\
Success-probability lower expression & \((0.82896891,\ 0.82896893)\)\\
Linear CF coefficient \(c\) & \((322.99298,\ 322.99299)\)\\
\bottomrule
\end{tabular}
\caption{Interval bounds used in the population and single-batch
proofs. The stated theorem constants are conservative consequences
of these enclosures.}
\label{tab:constants}
\end{table}

\section{One realized batch}
\label{app:batch}

Draw \(N=256\) iid scenes and eight independent cube nuisances per scene,
and fix this batch \(\mathcal B\). Define
\begin{equation}
 \widehat F_{\mathcal B}(h)=
 \frac1{NVk}\sum_{i,v}\norm{h_{iv}-\bar h_{i,G}}^2
 +\frac1{50}\E_\theta\widehat R_{\mathcal B,\theta}(h).
 \label{eq:singlebatch}
\end{equation}
The constraints remain population constraints \eqref{eq:whiten}.

\begin{theorem}[Uniform single-batch exclusion]
\label{thm:batch}
With probability at least \(0.828\) over \(\mathcal B\), every global
minimizer of \eqref{eq:singlebatch} has strictly positive population \(u\).
On the same event every minimizer in the linear Stiefel class satisfies
\(u(W)=\norm{B}_F^2/512>10^{-12}\), where \(W=[A,B]\).
\end{theorem}
\begin{proof}
Let \(q_s\) be the empirical scene frequencies and
\(a=q_1+q_2=K/256\), \(K\sim\mathrm{Bin}(256,1/2)\).
Every scene-only whitened encoder is \(Qr\), including rotations selected
after observing the batch. Its empirical self term is
\begin{align*}
 j_q(t)={}&q_1^2+q_2^2+2q_1q_2\psi_8(t)
 +(1-a)^2\psi_{2044}(t)\\
 &+(1-\psi_{2044}(t))\sum_{s=3}^{513}q_s^2
 +2a(1-a)\psi_{1026}(t).
\end{align*}
As the real sphere CF is at most one,
\(q_1^2+q_2^2\geq a^2/2\) and
\(\sum_{s\geq3}q_s^2\geq(1-a)^2/511\) imply
\[
 j_q(t)\geq H(t)a^2+T(t)(1-a)^2+2Z(t)a(1-a),
\]
where \(H=(1+\psi_8)/2\),
\(T=\psi_{2044}+(1-\psi_{2044})/511\), and \(Z=\psi_{1026}\).
Its target factor is exactly
\(x_q(t)=a\psi_3(t)+(1-a)\psi_{1021}(t)\).
Consequently its discrepancy is at least
\[
 d_{\min}(a)=\sum_j\omega_j\left[
 H_j a^2+T_j(1-a)^2+2Z_j a(1-a)
 -2g_j\{a\psi_3(t_j)+(1-a)\psi_{1021}(t_j)\}+g_j^2\right].
\]
This is a quadratic \(\alpha_0a^2+\beta_0a+\eta_0\).
The interval bounds in Appendix~\ref{app:sphere} give
\[
 \alpha_0>0,\quad
 2\alpha_0(117/256)+\beta_0>0,\quad
 \frac{2048}{50}d_{\min}(117/256)>1.5019>3/2.
\]
Therefore \(K\geq117\) implies
\(\widehat F_{\mathcal B}(Qr)>3/2\) for every \(Q\).
Empirical scene collisions are already included in \(q_s^2\);
no population bias formula is substituted into this lower bound.

For the cube, each sample-coordinate prediction contribution is
\[
 L=\frac18\sum_{v=1}^8
       \left(U_v-\frac{U_1+U_2}{2}\right)^2.
\]
Enumeration of its 256 sign patterns gives \(0\leq L\leq3\),
\(\E L=5/4\). There are \(256\cdot512\) independent contributions.
Hoeffding's inequality and EP nonnegativity with Markov's inequality give
\[
 \Pr(\widehat P_U>127/100)
 \leq e^{-2\cdot256\cdot512/(50^2\cdot9)},\qquad
 \Pr(\E_\theta\widehat R_U>11)\leq\E_{\mathcal B,\theta}\widehat R_U/11.
\]
The cube has \(\widehat F_U\leq149/100\) on their complementary events.
Table~\ref{tab:gap} gives \(\E\widehat R_U<1.053857\).
The exact scene-count tail and a union bound yield success probability
greater than
\begin{equation}
 1-2^{-256}\sum_{j=0}^{116}\binom{256}{j}
 -e^{-2\cdot256\cdot512/(50^2\cdot9)}
 -\frac{\E\widehat R_U}{11}>0.8289689182.
 \label{eq:prob}
\end{equation}
No independence between these three events is needed.

On this event a feasible cube lies below every scene-only encoder.
Compactness and continuity again give existence of minima.
Zero population \(u\) would make a minimum scene-only at every
positive-mass atom, a contradiction.

For the linear statement, polar decomposition gives a \(Q\) with
\(\norm{A-Q}_F\leq512u\), while \(\norm B_F=\sqrt{512u}\).
For every observation, using the maximum scene norm,
\[
 \norm{h_W-Qr}/\sqrt{512}
 \leq e(u):=\sqrt{512\cdot1021}\,u+\sqrt{512}\sqrt u.
\]
After direction averaging the empirical CF difference is at most
\(|t|e(u)\). The square discrepancies from a target of magnitude at
most one differ by at most four times the CF difference, hence
\[
 \frac1{50}\left|\E_\theta\widehat R_W
      -\E_\theta\widehat R_{Qr}\right|
 \leq c\,e(u),\qquad
 c=\frac{4\cdot2048}{50}\sum_j\omega_j|t_j|<323.
\]
Nonnegative prediction gives
\(\widehat F(Qr)\leq\widehat F(W)+ce(u)\).
On the good event every optimum has value at most \(1.49\), whereas
\(Qr>1.5\), so \(ce(u)>0.01\).
Substitution gives \(ce(10^{-12})<0.007309\).
Monotonicity of \(e\) proves \(u>10^{-12}\).
\end{proof}

The single-batch theorem covers rotations chosen after observing the
batch because the sphere integral is exactly rotation invariant.
The linear bound is conservative; the event argument establishes
exclusion of the entire scene-only class.

\section{Finite-width encoder realization}
\label{app:realization}
Let the injective observation values be \(\tau=(-3,-1,1,3)\).
The model
\[
 h(\tau)=b+c_1(\tau+3)_++c_2(\tau+1)_++c_3(\tau-1)_+
\]
has evaluation matrix
\[
 \begin{pmatrix}
 1&0&0&0\\
 1&2&0&0\\
 1&4&2&0\\
 1&6&4&2
 \end{pmatrix}
\]
for parameters \((b,c_1,c_2,c_3)\). Its determinant is 8.
Thus every four-state scalar encoder has a unique realization in
this fixed-feature ReLU class. The scalar objective-level results
apply directly to this parameterization.

\section{A noisy-readout inequality}
\label{app:information}
Let \(C=(S,U)\) be finite, \(Y\) finite-valued, and \(h(C)\)
deterministic with finite variance. Set
\(Z_\sigma=h(C)+\sigma\xi\), where \(\sigma>0\) and
\(\xi\sim\mathcal N(0,1)\) is independent of \((C,Y)\).
Write \(u=\E\Var(h\mid S)\) and \(\Delta=I(Y;U\mid S)\).
For unrestricted measurable probability decoders,
\begin{equation}
 H(Y\mid Z_\sigma)-H(Y\mid S,U)
 \geq\max\left\{0,\Delta-\frac12\log(1+u/\sigma^2)\right\}.
 \label{eq:infoineq}
\end{equation}
To prove it, provide \(S\) additionally to the decoder:
\[
 H(Y\mid Z_\sigma)-H(Y\mid S,U)
 \geq\Delta-I(Y;Z_\sigma\mid S)
 \geq\Delta-I(U;Z_\sigma\mid S).
\]
The second inequality is conditional data processing.
For each \(S=s\), Gaussian maximal differential entropy at fixed
variance gives
\[
 I(U;Z_\sigma\mid S=s)
 \leq\frac12\log\left(1+
 \frac{\Var(h\mid S=s)}{\sigma^2}\right).
\]
Averaging and concavity of the logarithm yield
\eqref{eq:infoineq}, and data processing gives its nonnegative
floor \citep{cover}. The bound also applies when \(S\) is
available to the decoder.

\section{Experimental details}
\label{app:protocol}
\subsection{Encoder training and paired data}
The encoder uses the fixed-feature ReLU parameterization in
Appendix~\ref{app:realization}. Prediction training uses the eight-view
center loss with \(G=2,N=256\). The regularized branches use the
pooled quadrature in \eqref{eq:poolednodes} with coefficient \(0.02\).
Population whitening centers and rescales the four outputs before
evaluating the whitened branches. NTP jointly trains a whitened encoder
and a scalar linear logit head against the conditional token
probabilities in \eqref{eq:token}. All encoders use 2,000 Adam updates
with learning rate \(10^{-3}\) and five random seeds.
Positive-view batches are shared within an encoder seed. The three
JEPA branches respectively use whitening plus prediction, soft pooled
EP, and whitening plus pooled EP. NTP uses a binary linear token head.
Readout fitting uses sampled Bernoulli labels rather than the
conditional token expectations used during NTP encoder training.

\subsection{Standardization and label budgets}
Each frozen encoder is centered and divided by its four-state population
standard deviation. The readout input is
\(Z=z+\sigma\xi\), \(\xi\sim\mathcal N(0,1)\),
with \(\sigma\in\{0,0.01\}\) applied consistently to training,
validation, and testing. Training budgets
\(n=32,128,512,2048\) use validation budgets \(32,32,128,512\).
State overlap between splits is allowed; observations and sampled labels
are independent across pools. All methods, encoder seeds, heads, and
noise conditions share the state, token, and Gaussian draws within a
repeat. The test set selects no head, hyperparameter, or checkpoint.

For each data repeat, independent training, validation, and test
pools have 2,048, 512, and 16,384 samples. Each pool draws uniform
state indices, uniform token variables, and standard Gaussian noise.
Labels are Bernoulli samples from \eqref{eq:token}.
Data are shared across methods and encoder seeds within a repeat;
smaller budgets use prefixes of the same pools.
The main \(b=2\) experiment crosses five encoder seeds, three data
repeats, four training budgets, two noise levels, six representations
and two head architectures, giving 1,440 fits.
The \(b=0\) diagnostic uses one encoder seed, two data repeats, two
budgets, two noise levels, two representations and linear heads,
giving 32 additional fits.

The controls \(S\) and \((S+2U)/\sqrt5\) are deterministic
representations. Repeating them across encoder seeds matches head
initialization and data pairing without introducing additional
learned encoders.

\subsection{Readout fitting}
The linear head is \(\ell(Z)=wZ+b_0\), initialized at zero.
L-BFGS-B minimizes mean sampled-label cross entropy plus
\(10^{-4}(w^2+b_0^2)/2\), with maximum 2,000 iterations,
\texttt{gtol}\(=10^{-8}\), and \texttt{ftol}\(=10^{-12}\).
The final optimizer parameters define the fitted head.

The MLP is
\(\ell(Z)=w_2^\top\tanh(w_1Z+b_1)+b_2\), with 49 parameters.
For each linear layer, both weights and biases are initialized
uniformly on
\([-1/\sqrt{\mathrm{fan\_in}},1/\sqrt{\mathrm{fan\_in}}]\).
Initializations are shared across methods at a fixed encoder seed
and data repeat.
Full-batch Adam uses learning rate \(0.01\),
\((\beta_1,\beta_2)=(0.9,0.999)\),
\(\epsilon_{\rm Adam}=10^{-8}\), and coupled L2 gradient \(10^{-4}\)
on all parameters. Validation NLL is evaluated at update zero and
every ten updates. Training stops after twenty evaluations without
an improvement exceeding \(10^{-6}\) over the last significant
improvement, or at 1,000 updates. The selected checkpoint is the
earliest strictly lowest validation NLL.

All fits and quadrature calculations use float64 with NumPy 2.0.2
and SciPy 1.13.1. At zero noise, repeated exact observation values
are aggregated using sample counts and sampled-label means,
preserving the full-batch loss and gradient.
All 752 linear fits met the solver's convergence criterion.
Of 720 MLP fits, 187 reached the update cap and 533 stopped by
validation patience. The MLP comparison therefore evaluates finite
optimization trajectories.

\subsection{Population risk and uncertainty}
For \(\sigma>0\), the posterior state weights at observation \(z\)
are proportional to
\(\exp(-(z-z_i)^2/(2\sigma^2))\).
Their weighted token probabilities determine \(R_{\rm obs}\).
Population expectations are sums of four Gaussian integrals,
evaluated with probabilists' Hermite quadrature and log-sum-exp
stabilization. At \(\sigma=0\), exactly equal float64 values are
grouped before computing conditional token entropy.

Each of 2,000 bootstrap draws independently resamples the five
encoder-seed indices and three shared-data-repeat indices with
replacement, then averages the crossed paired differences.
Quantiles at 0.025 and 0.975 form the empirical intervals.
This procedure preserves the two sources of replication rather
than treating the 15 cells as independent encoder seeds.

\section{Additional readout results}
\label{app:results}
\begin{table}[ht]
\centering
\begin{tabular}{rrrrr}
\toprule
Training \(n\) & Linear, \(\sigma=0\) & Linear, 0.01 &
MLP16, \(\sigma=0\) & MLP16, 0.01\\
\midrule
32 & 0.271382 & 0.271181 & 0.229097 & 0.229478\\
128 & 0.290776 & 0.290745 & 0.237620 & 0.237916\\
512 & 0.287282 & 0.287226 & 0.228746 & 0.229160\\
2048 & 0.286032 & 0.285984 & 0.226737 & 0.226463\\
\bottomrule
\end{tabular}
\caption{Mean whitened-prediction JEPA minus NTP sampled-token
test NLL on \(b=2\). Positive means lower risk for NTP. Aggregate
MLP differences coexist with variation across seeds and paired cells discussed
in Section~\ref{sec:readout}.}
\end{table}

\begin{table}[ht]
\centering
\begin{tabular}{cr}
\toprule
Encoder seed & Primary linear difference, mean of three repeats\\
\midrule
0 & 0.285893\\
1 & 0.285844\\
2 & 0.285899\\
3 & 0.292616\\
4 & 0.285879\\
\bottomrule
\end{tabular}
\caption{Seed means for the primary linear comparison at
\(n=512,\sigma=0.01\).}
\end{table}

\subsection{Main comparison and controls}
The prespecified primary comparison uses \(b=2,n=512\),
128 validation labels, \(\sigma=0.01\), and the linear head.
The aggregate difference is JEPA minus NTP sampled-token test NLL.
Confidence intervals independently resample the five encoder-seed and
three shared-data-repeat axes while preserving the method pairing;
2,000 bootstrap draws are used.

The linear primary difference has a 95\% two-axis paired bootstrap
interval \([0.280889,0.293062]\). All five seed means and all 15 paired
seed/repeat cells are positive. The MLP secondary difference has interval
\([0.115088,0.291289]\); seed 3's mean difference is only \(0.001872\),
showing variation in effect size across encoder seeds.
On the same test states and noise, conditional-expectation NLL gives a
primary difference \(0.284142\), interval \([0.281504,0.287139]\).
For the fixed fitted heads, the paired test Monte Carlo standard error
is \(0.002677\); it measures a different uncertainty from the bootstrap.

\begin{figurehere}
\centering
\includegraphics[width=\linewidth]{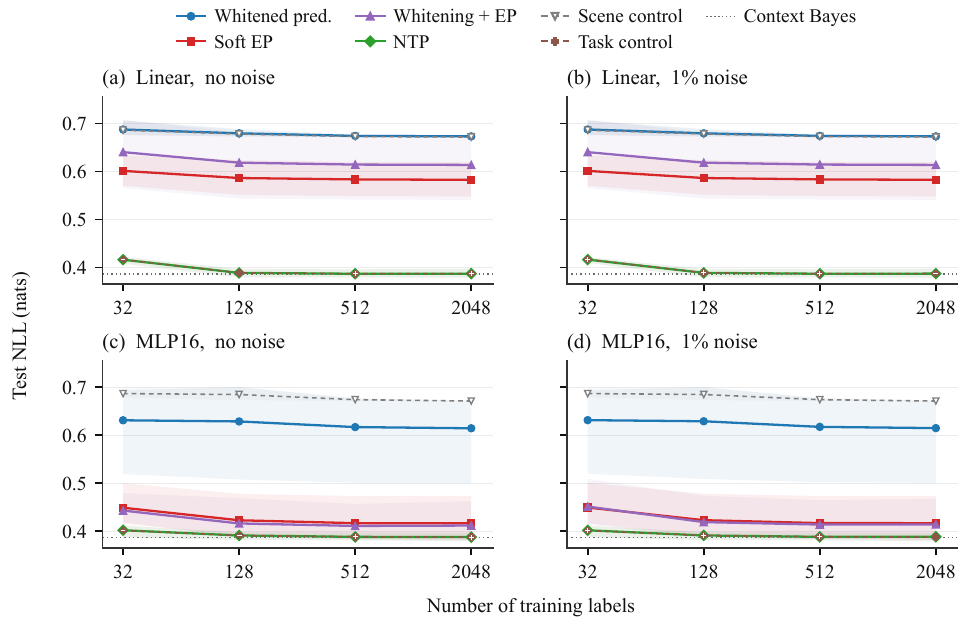}
\caption{Readout across budgets and noise.}
\label{fig:curves}
\end{figurehere}

Figure~\ref{fig:curves} shows every budget, both noise settings, both heads,
and both soft JEPA branches on the constructed \(b=2\) task.
All six representations use common axes and a shared legend.
Bands are 95\% two-axis bootstrap intervals over five encoder seeds
and three shared-data repeats. Dashed controls use fixed representations
with the same decoder and paired data; distinct markers distinguish
overlapping curves. The dotted line is the complete-context Bayes risk,
\(0.386534\) nats/token. The lower-panel scene control illustrates that
equal clean alignment does not determine nonlinear prediction risk.
Appendix~\ref{app:results} lists all eight
budget/noise differences for the primary representation pair.
The linear ordering is positive in every cell. For noiseless MLP16 at
\(n=512\), seed 3's mean difference is \(-0.000281\); at \(n=2048\)
only four seed means are positive. The positive aggregate MLP differences
therefore coexist with variation in the ranking across encoder seeds
and label budgets.

At \(n=512,\sigma=0.01\), soft EP and whitening-plus-EP MLP test NLLs
are \(0.417081\) and \(0.414023\). Their differences from NTP are
\(0.028651\) and \(0.025593\), with respective intervals
\([-0.000919,0.085200]\) and \([-0.001071,0.077731]\).
Both intervals cross zero and include seed/cell reversals. The primary
whitened-prediction branch has a larger aggregate gap from NTP than
either soft Gaussian branch with a matched nonlinear head.

For \(b=0\), the diagnostic uses seed 0 and two repeats with linear heads,
for 32 total fits. The whitened JEPA minus NTP differences at
\((n,\sigma)=(32,0),(32,0.01),(512,0),(512,0.01)\) are respectively
\(-0.017188,-0.017424,-0.016543,-0.016546\).
This control reverses the main ordering. With \(b=0\), both ideal
representations can attain the scene-based optimum; the finite-step
NTP encoder can nevertheless have higher risk. This diagnostic uses
one encoder seed.

\subsection{Information loss versus the remaining head gap}
We evaluate both terms in the risk decomposition \eqref{eq:decomp}.
The remaining head gap jointly contains capacity, finite training,
regularization, and optimization effects; it is not solely sample error.
For any representation channel independent of \(Y\) given the context,
\(R_{\rm obs}-R_{\rm context}=I(Y;S,U\mid Z)\) in nats.
This follows from the conditional-entropy characterization of optimal
log loss \citep{cover}.

For \(\sigma>0\), \(Z\) follows a four-component Gaussian mixture and
its posterior token probability can be evaluated directly.
For \(\sigma=0\), exact equality classes of the executed float64 values
define the observation. No tolerance-based merging is used.
The complete-context Bayes risk for \(b=2\) is \(0.386534\) nats/token.
The following quantities use 128-node numerical Hermite quadrature.

\begin{tablehere}
\centering\small
\begin{tabular}{lrrr}
\toprule
Representation & Observation loss & Linear head gap & MLP16 head gap\\
\midrule
Whitened prediction & 0.225499 & 0.059349 & 0.003077\\
Soft pooled EP & 0.027999 & 0.165827 & 0.002191\\
Whitening plus pooled EP & 0.019571 & 0.207490 & 0.007651\\
NTP & \(\approx0\) & 0.000365 & 0.002184\\
Scene-only control & 0.281874 & 0.001552 & 0.002657\\
Task-sufficient control & \(\approx0\) & 0.000301 & 0.002358\\
\bottomrule
\end{tabular}
\caption{Population risk decomposition at \(n=512,\sigma=0.01\),
averaged over seeds and repeats and evaluated by numerical quadrature.
``Approximately zero'' denotes values below numerical resolution.}
\label{tab:decomp}
\end{tablehere}

At \(n=2048\) the primary JEPA MLP head gap falls to \(0.000996\), while
the average observation loss remains \(0.225499\).
Section~\ref{sec:readout} distinguishes exact-value injectivity from
information available through a noisy channel.

The 64- and 128-node head-risk estimates differ by at most
\(4.44\times10^{-16}\), and observed Bayes risks by at most
\(2.95\times10^{-15}\), indicating numerical agreement.

\clearpage
\section{Complete rational partition certificate}
\label{app:partition}
Table~\ref{tab:scalar_partition} supplies every rectangle used in the
global phase coverage argument of Appendix~\ref{app:cert}.
For a scene-only vector \((b,b,c,c)\), the phase coordinates satisfy
\(b=8\pi u/5\) and \(c=8\pi v/5\). The rectangles cover \([-1,1]^2\)
and have disjoint interiors. At node \(t_j=5j/8\), the two-atom
characteristic function is \((e^{ij\pi u}+e^{ij\pi v})/2\).

All interval endpoints are exact rational numbers.
Each listed lower bound applies to \(D_{\rm gr}\) throughout the
corresponding rectangle and is rounded downward to twelve decimal places
from the outward interval enclosure. Every listed bound exceeds \(1/50\);
the smallest exceeds \(0.0200111\). The conversion from discrepancy
bounds to objective bounds is given in Appendix~\ref{app:scalar}.

\begingroup
\small
\setlength{\tabcolsep}{9pt}
\renewcommand{\arraystretch}{1.02}
\begin{longtable}{rrrrrr}
\caption{Rational partition certificate.}
\label{tab:scalar_partition}\\
\toprule
No. & \(u_-\) & \(u_+\) & \(v_-\) & \(v_+\) & Lower bound on \(D_{\rm gr}\)\\
\midrule
\endfirsthead
\multicolumn{6}{c}{Table \thetable: continued}\\
\toprule
No. & \(u_-\) & \(u_+\) & \(v_-\) & \(v_+\) & Lower bound on \(D_{\rm gr}\)\\
\midrule
\endhead
\midrule
\multicolumn{6}{r}{Continued on the next page}\\
\endfoot
\bottomrule
\endlastfoot
1 & \(1/2\) & \(1\) & \(0\) & \(1\) & \(0.106992955312\)\\
2 & \(0\) & \(1/2\) & \(1/2\) & \(1\) & \(0.106992955312\)\\
3 & \(1/4\) & \(1/2\) & \(0\) & \(1/2\) & \(0.128527744124\)\\
4 & \(0\) & \(1/4\) & \(1/4\) & \(1/2\) & \(0.128527744124\)\\
5 & \(1/8\) & \(1/4\) & \(0\) & \(1/4\) & \(0.136122246193\)\\
6 & \(0\) & \(1/8\) & \(1/8\) & \(1/4\) & \(0.136122246193\)\\
7 & \(0\) & \(1/8\) & \(0\) & \(1/8\) & \(0.055640786885\)\\
8 & \(1/2\) & \(1\) & \(-1\) & \(0\) & \(0.106992955312\)\\
9 & \(3/8\) & \(1/2\) & \(-1/4\) & \(0\) & \(0.093169239042\)\\
10 & \(1/4\) & \(3/8\) & \(-1/8\) & \(0\) & \(0.027547335016\)\\
11 & \(1/4\) & \(3/8\) & \(-1/4\) & \(-1/8\) & \(0.030515503813\)\\
12 & \(1/4\) & \(1/2\) & \(-1/2\) & \(-1/4\) & \(0.140076106025\)\\
13 & \(3/16\) & \(1/4\) & \(-1/16\) & \(0\) & \(0.077076164041\)\\
14 & \(7/32\) & \(1/4\) & \(-1/8\) & \(-1/16\) & \(0.033446519418\)\\
15 & \(3/16\) & \(7/32\) & \(-3/32\) & \(-1/16\) & \(0.043766160118\)\\
16 & \(3/16\) & \(7/32\) & \(-1/8\) & \(-3/32\) & \(0.021659582602\)\\
17 & \(1/8\) & \(3/16\) & \(-1/16\) & \(0\) & \(0.065032436996\)\\
18 & \(5/32\) & \(3/16\) & \(-3/32\) & \(-1/16\) & \(0.040633798522\)\\
19 & \(11/64\) & \(3/16\) & \(-1/8\) & \(-3/32\) & \(0.020246456354\)\\
20 & \(5/32\) & \(11/64\) & \(-1/8\) & \(-3/32\) & \(0.021113680162\)\\
21 & \(1/8\) & \(5/32\) & \(-1/8\) & \(-1/16\) & \(0.024897899225\)\\
22 & \(7/32\) & \(1/4\) & \(-3/16\) & \(-1/8\) & \(0.031452014370\)\\
23 & \(13/64\) & \(7/32\) & \(-5/32\) & \(-1/8\) & \(0.030642300628\)\\
24 & \(3/16\) & \(13/64\) & \(-9/64\) & \(-1/8\) & \(0.025738318213\)\\
25 & \(3/16\) & \(13/64\) & \(-5/32\) & \(-9/64\) & \(0.029651717957\)\\
26 & \(3/16\) & \(7/32\) & \(-3/16\) & \(-5/32\) & \(0.037385192550\)\\
27 & \(3/16\) & \(1/4\) & \(-1/4\) & \(-3/16\) & \(0.068158184816\)\\
28 & \(23/128\) & \(3/16\) & \(-9/64\) & \(-1/8\) & \(0.022940307702\)\\
29 & \(11/64\) & \(23/128\) & \(-9/64\) & \(-1/8\) & \(0.020753402532\)\\
30 & \(23/128\) & \(3/16\) & \(-5/32\) & \(-9/64\) & \(0.024656824567\)\\
31 & \(11/64\) & \(23/128\) & \(-5/32\) & \(-9/64\) & \(0.020763826933\)\\
32 & \(21/128\) & \(11/64\) & \(-9/64\) & \(-1/8\) & \(0.020011105452\)\\
33 & \(5/32\) & \(21/128\) & \(-9/64\) & \(-1/8\) & \(0.020328489150\)\\
34 & \(21/128\) & \(11/64\) & \(-19/128\) & \(-9/64\) & \(0.021460556323\)\\
35 & \(21/128\) & \(11/64\) & \(-5/32\) & \(-19/128\) & \(0.022241800413\)\\
36 & \(5/32\) & \(21/128\) & \(-19/128\) & \(-9/64\) & \(0.020919833707\)\\
37 & \(5/32\) & \(21/128\) & \(-5/32\) & \(-19/128\) & \(0.020854210365\)\\
38 & \(11/64\) & \(3/16\) & \(-3/16\) & \(-5/32\) & \(0.027006432780\)\\
39 & \(21/128\) & \(11/64\) & \(-11/64\) & \(-5/32\) & \(0.023670830689\)\\
40 & \(5/32\) & \(21/128\) & \(-11/64\) & \(-5/32\) & \(0.020541080218\)\\
41 & \(5/32\) & \(11/64\) & \(-3/16\) & \(-11/64\) & \(0.027496222894\)\\
42 & \(9/64\) & \(5/32\) & \(-9/64\) & \(-1/8\) & \(0.020071320341\)\\
43 & \(19/128\) & \(5/32\) & \(-19/128\) & \(-9/64\) & \(0.021805057753\)\\
44 & \(19/128\) & \(5/32\) & \(-5/32\) & \(-19/128\) & \(0.020887480015\)\\
45 & \(9/64\) & \(19/128\) & \(-19/128\) & \(-9/64\) & \(0.023542148758\)\\
46 & \(9/64\) & \(19/128\) & \(-5/32\) & \(-19/128\) & \(0.021805057753\)\\
47 & \(1/8\) & \(9/64\) & \(-9/64\) & \(-1/8\) & \(0.027906682015\)\\
48 & \(1/8\) & \(9/64\) & \(-5/32\) & \(-9/64\) & \(0.020071320341\)\\
49 & \(19/128\) & \(5/32\) & \(-21/128\) & \(-5/32\) & \(0.020854210365\)\\
50 & \(19/128\) & \(5/32\) & \(-11/64\) & \(-21/128\) & \(0.022241800413\)\\
51 & \(9/64\) & \(19/128\) & \(-21/128\) & \(-5/32\) & \(0.020919833707\)\\
52 & \(9/64\) & \(19/128\) & \(-11/64\) & \(-21/128\) & \(0.021460556323\)\\
53 & \(19/128\) & \(5/32\) & \(-3/16\) & \(-11/64\) & \(0.024135790080\)\\
54 & \(9/64\) & \(19/128\) & \(-3/16\) & \(-11/64\) & \(0.021895720944\)\\
55 & \(17/128\) & \(9/64\) & \(-21/128\) & \(-5/32\) & \(0.022463125432\)\\
56 & \(17/128\) & \(9/64\) & \(-11/64\) & \(-21/128\) & \(0.022424936622\)\\
57 & \(1/8\) & \(17/128\) & \(-11/64\) & \(-5/32\) & \(0.021324454351\)\\
58 & \(17/128\) & \(9/64\) & \(-3/16\) & \(-11/64\) & \(0.021428004784\)\\
59 & \(1/8\) & \(17/128\) & \(-3/16\) & \(-11/64\) & \(0.022754701424\)\\
60 & \(5/32\) & \(3/16\) & \(-1/4\) & \(-3/16\) & \(0.035094765240\)\\
61 & \(9/64\) & \(5/32\) & \(-7/32\) & \(-3/16\) & \(0.028885860212\)\\
62 & \(1/8\) & \(9/64\) & \(-7/32\) & \(-3/16\) & \(0.024663619036\)\\
63 & \(1/8\) & \(5/32\) & \(-1/4\) & \(-7/32\) & \(0.043693954385\)\\
64 & \(0\) & \(1/8\) & \(-1/8\) & \(0\) & \(0.055640786885\)\\
65 & \(3/32\) & \(1/8\) & \(-5/32\) & \(-1/8\) & \(0.025338755960\)\\
66 & \(7/64\) & \(1/8\) & \(-11/64\) & \(-5/32\) & \(0.023253273071\)\\
67 & \(7/64\) & \(1/8\) & \(-3/16\) & \(-11/64\) & \(0.023561156316\)\\
68 & \(3/32\) & \(7/64\) & \(-3/16\) & \(-5/32\) & \(0.025521547372\)\\
69 & \(1/16\) & \(3/32\) & \(-3/16\) & \(-1/8\) & \(0.025260401259\)\\
70 & \(3/32\) & \(1/8\) & \(-7/32\) & \(-3/16\) & \(0.021659582602\)\\
71 & \(3/32\) & \(1/8\) & \(-1/4\) & \(-7/32\) & \(0.043342332898\)\\
72 & \(1/16\) & \(3/32\) & \(-1/4\) & \(-3/16\) & \(0.039247869050\)\\
73 & \(0\) & \(1/16\) & \(-1/4\) & \(-1/8\) & \(0.025821035502\)\\
74 & \(1/8\) & \(1/4\) & \(-3/8\) & \(-1/4\) & \(0.030515503813\)\\
75 & \(1/8\) & \(1/4\) & \(-1/2\) & \(-3/8\) & \(0.225500317491\)\\
76 & \(0\) & \(1/8\) & \(-1/2\) & \(-1/4\) & \(0.027055221046\)\\
77 & \(0\) & \(1/2\) & \(-1\) & \(-1/2\) & \(0.106992955312\)\\
78 & \(-1/2\) & \(0\) & \(1/2\) & \(1\) & \(0.106992955312\)\\
79 & \(-1/8\) & \(0\) & \(1/4\) & \(1/2\) & \(0.027055221046\)\\
80 & \(-1/4\) & \(-1/8\) & \(3/8\) & \(1/2\) & \(0.225500317491\)\\
81 & \(-1/4\) & \(-1/8\) & \(1/4\) & \(3/8\) & \(0.030515503813\)\\
82 & \(-1/16\) & \(0\) & \(1/8\) & \(1/4\) & \(0.025821035502\)\\
83 & \(-3/32\) & \(-1/16\) & \(3/16\) & \(1/4\) & \(0.039247869050\)\\
84 & \(-1/8\) & \(-3/32\) & \(7/32\) & \(1/4\) & \(0.043342332898\)\\
85 & \(-1/8\) & \(-3/32\) & \(3/16\) & \(7/32\) & \(0.021659582602\)\\
86 & \(-3/32\) & \(-1/16\) & \(1/8\) & \(3/16\) & \(0.025260401259\)\\
87 & \(-7/64\) & \(-3/32\) & \(5/32\) & \(3/16\) & \(0.025521547372\)\\
88 & \(-1/8\) & \(-7/64\) & \(11/64\) & \(3/16\) & \(0.023561156316\)\\
89 & \(-1/8\) & \(-7/64\) & \(5/32\) & \(11/64\) & \(0.023253273071\)\\
90 & \(-1/8\) & \(-3/32\) & \(1/8\) & \(5/32\) & \(0.025338755960\)\\
91 & \(-1/8\) & \(0\) & \(0\) & \(1/8\) & \(0.055640786885\)\\
92 & \(-5/32\) & \(-1/8\) & \(7/32\) & \(1/4\) & \(0.043693954385\)\\
93 & \(-9/64\) & \(-1/8\) & \(3/16\) & \(7/32\) & \(0.024663619036\)\\
94 & \(-5/32\) & \(-9/64\) & \(3/16\) & \(7/32\) & \(0.028885860212\)\\
95 & \(-3/16\) & \(-5/32\) & \(3/16\) & \(1/4\) & \(0.035094765240\)\\
96 & \(-17/128\) & \(-1/8\) & \(11/64\) & \(3/16\) & \(0.022754701424\)\\
97 & \(-9/64\) & \(-17/128\) & \(11/64\) & \(3/16\) & \(0.021428004784\)\\
98 & \(-17/128\) & \(-1/8\) & \(5/32\) & \(11/64\) & \(0.021324454351\)\\
99 & \(-9/64\) & \(-17/128\) & \(21/128\) & \(11/64\) & \(0.022424936622\)\\
100 & \(-9/64\) & \(-17/128\) & \(5/32\) & \(21/128\) & \(0.022463125432\)\\
101 & \(-19/128\) & \(-9/64\) & \(11/64\) & \(3/16\) & \(0.021895720944\)\\
102 & \(-5/32\) & \(-19/128\) & \(11/64\) & \(3/16\) & \(0.024135790080\)\\
103 & \(-19/128\) & \(-9/64\) & \(21/128\) & \(11/64\) & \(0.021460556323\)\\
104 & \(-19/128\) & \(-9/64\) & \(5/32\) & \(21/128\) & \(0.020919833707\)\\
105 & \(-5/32\) & \(-19/128\) & \(21/128\) & \(11/64\) & \(0.022241800413\)\\
106 & \(-5/32\) & \(-19/128\) & \(5/32\) & \(21/128\) & \(0.020854210365\)\\
107 & \(-9/64\) & \(-1/8\) & \(9/64\) & \(5/32\) & \(0.020071320341\)\\
108 & \(-9/64\) & \(-1/8\) & \(1/8\) & \(9/64\) & \(0.027906682015\)\\
109 & \(-19/128\) & \(-9/64\) & \(19/128\) & \(5/32\) & \(0.021805057753\)\\
110 & \(-19/128\) & \(-9/64\) & \(9/64\) & \(19/128\) & \(0.023542148758\)\\
111 & \(-5/32\) & \(-19/128\) & \(19/128\) & \(5/32\) & \(0.020887480015\)\\
112 & \(-5/32\) & \(-19/128\) & \(9/64\) & \(19/128\) & \(0.021805057753\)\\
113 & \(-5/32\) & \(-9/64\) & \(1/8\) & \(9/64\) & \(0.020071320341\)\\
114 & \(-11/64\) & \(-5/32\) & \(11/64\) & \(3/16\) & \(0.027496222894\)\\
115 & \(-21/128\) & \(-5/32\) & \(5/32\) & \(11/64\) & \(0.020541080218\)\\
116 & \(-11/64\) & \(-21/128\) & \(5/32\) & \(11/64\) & \(0.023670830689\)\\
117 & \(-3/16\) & \(-11/64\) & \(5/32\) & \(3/16\) & \(0.027006432780\)\\
118 & \(-21/128\) & \(-5/32\) & \(19/128\) & \(5/32\) & \(0.020854210365\)\\
119 & \(-21/128\) & \(-5/32\) & \(9/64\) & \(19/128\) & \(0.020919833707\)\\
120 & \(-11/64\) & \(-21/128\) & \(19/128\) & \(5/32\) & \(0.022241800413\)\\
121 & \(-11/64\) & \(-21/128\) & \(9/64\) & \(19/128\) & \(0.021460556323\)\\
122 & \(-21/128\) & \(-5/32\) & \(1/8\) & \(9/64\) & \(0.020328489150\)\\
123 & \(-11/64\) & \(-21/128\) & \(1/8\) & \(9/64\) & \(0.020011105452\)\\
124 & \(-23/128\) & \(-11/64\) & \(9/64\) & \(5/32\) & \(0.020763826933\)\\
125 & \(-3/16\) & \(-23/128\) & \(9/64\) & \(5/32\) & \(0.024656824567\)\\
126 & \(-23/128\) & \(-11/64\) & \(1/8\) & \(9/64\) & \(0.020753402532\)\\
127 & \(-3/16\) & \(-23/128\) & \(1/8\) & \(9/64\) & \(0.022940307702\)\\
128 & \(-1/4\) & \(-3/16\) & \(3/16\) & \(1/4\) & \(0.068158184816\)\\
129 & \(-7/32\) & \(-3/16\) & \(5/32\) & \(3/16\) & \(0.037385192550\)\\
130 & \(-13/64\) & \(-3/16\) & \(9/64\) & \(5/32\) & \(0.029651717957\)\\
131 & \(-13/64\) & \(-3/16\) & \(1/8\) & \(9/64\) & \(0.025738318213\)\\
132 & \(-7/32\) & \(-13/64\) & \(1/8\) & \(5/32\) & \(0.030642300628\)\\
133 & \(-1/4\) & \(-7/32\) & \(1/8\) & \(3/16\) & \(0.031452014370\)\\
134 & \(-5/32\) & \(-1/8\) & \(1/16\) & \(1/8\) & \(0.024897899225\)\\
135 & \(-11/64\) & \(-5/32\) & \(3/32\) & \(1/8\) & \(0.021113680162\)\\
136 & \(-3/16\) & \(-11/64\) & \(3/32\) & \(1/8\) & \(0.020246456354\)\\
137 & \(-3/16\) & \(-5/32\) & \(1/16\) & \(3/32\) & \(0.040633798522\)\\
138 & \(-3/16\) & \(-1/8\) & \(0\) & \(1/16\) & \(0.065032436996\)\\
139 & \(-7/32\) & \(-3/16\) & \(3/32\) & \(1/8\) & \(0.021659582602\)\\
140 & \(-7/32\) & \(-3/16\) & \(1/16\) & \(3/32\) & \(0.043766160118\)\\
141 & \(-1/4\) & \(-7/32\) & \(1/16\) & \(1/8\) & \(0.033446519418\)\\
142 & \(-1/4\) & \(-3/16\) & \(0\) & \(1/16\) & \(0.077076164041\)\\
143 & \(-1/2\) & \(-1/4\) & \(1/4\) & \(1/2\) & \(0.140076106025\)\\
144 & \(-3/8\) & \(-1/4\) & \(1/8\) & \(1/4\) & \(0.030515503813\)\\
145 & \(-3/8\) & \(-1/4\) & \(0\) & \(1/8\) & \(0.027547335016\)\\
146 & \(-1/2\) & \(-3/8\) & \(0\) & \(1/4\) & \(0.093169239042\)\\
147 & \(-1\) & \(-1/2\) & \(0\) & \(1\) & \(0.106992955312\)\\
148 & \(-1/8\) & \(0\) & \(-1/8\) & \(0\) & \(0.055640786885\)\\
149 & \(-1/8\) & \(0\) & \(-1/4\) & \(-1/8\) & \(0.136122246193\)\\
150 & \(-1/4\) & \(-1/8\) & \(-1/4\) & \(0\) & \(0.136122246193\)\\
151 & \(-1/4\) & \(0\) & \(-1/2\) & \(-1/4\) & \(0.128527744124\)\\
152 & \(-1/2\) & \(-1/4\) & \(-1/2\) & \(0\) & \(0.128527744124\)\\
153 & \(-1/2\) & \(0\) & \(-1\) & \(-1/2\) & \(0.106992955312\)\\
154 & \(-1\) & \(-1/2\) & \(-1\) & \(0\) & \(0.106992955312\)\\

\end{longtable}
\endgroup

\paragraph{Use of generative AI tools.}
Generative AI tools accessed through TRAE assisted with mathematical
exploration and derivations, manuscript drafting, and analysis and
verification code. The author is responsible for all claims, proofs,
and results.

\end{document}